\documentclass{article} %

\usepackage{microtype}
\usepackage{graphicx}
\usepackage{subfigure}
\usepackage{booktabs} %

\usepackage{amsmath}
\usepackage{amsfonts}
\usepackage{bm}

\usepackage{hyperref}

\usepackage{dblfloatfix}

\usepackage[accepted]{icml2024}

\def\eqref#1{equation~\ref{#1}}

\def\1{\bm{1}}

\DeclareMathAlphabet{\mathsfit}{\encodingdefault}{\sfdefault}{m}{sl}
\SetMathAlphabet{\mathsfit}{bold}{\encodingdefault}{\sfdefault}{bx}{n}

\renewcommand{\paragraph}[1]{ \noindent \textbf{#1}}

\usepackage{url}
\usepackage{enumitem}
\usepackage{graphicx}
\usepackage{multirow}
\usepackage{booktabs}
\usepackage{adjustbox}  %
\usepackage{dblfloatfix}

\usepackage{graphicx}
\usepackage{amssymb}
\usepackage{minitoc}

\usepackage{natbib}
\usepackage{arydshln}

\setlist[enumerate]{itemsep=0.1pt, parsep=0.1pt, leftmargin=0.5cm}

\def\algname {SPADE}
\RequirePackage{algorithm}
\RequirePackage{algorithmic}

\icmltitlerunning{\algname{}: Sparsity-Guided Debugging for Deep Neural Networks}

\begin{document}

\twocolumn[
\icmltitle{\algname{}: Sparsity-Guided Debugging for Deep Neural Networks}

\icmlsetsymbol{equal}{*}

\begin{icmlauthorlist}
\icmlauthor{Arshia Soltani Moakhar}{equal,ista}
\icmlauthor{Eugenia Iofinova}{equal,ista}
\icmlauthor{Elias Frantar}{ista}
\icmlauthor{Dan Alistarh}{ista,nm}
\end{icmlauthorlist}

\icmlaffiliation{ista}{Institute of Science and Technology Austria (ISTA)}
\icmlaffiliation{nm}{NeuralMagic}

\icmlcorrespondingauthor{Eugenia Iofinova}{eugenia.iofinova@ista.ac.at}
\icmlcorrespondingauthor{Dan Alistarh}{dan.alistarh@ista.ac.at}

\icmlkeywords{Machine Learning, ICML}

\vskip 0.3in
]

\printAffiliationsAndNotice{\icmlEqualContribution} %

\doparttoc %
\faketableofcontents %

\vspace{-1em}
\begin{abstract}
It is known that sparsity can improve interpretability for deep neural networks. However, existing methods in the area either require networks that are pre-trained with sparsity constraints, or impose sparsity after the fact, altering the network's general behavior. In this paper, we demonstrate, for the first time, that sparsity can instead be incorporated into the interpretation process itself, as a sample-specific preprocessing step. Unlike previous work, this approach, which we call \algname{}, does not place constraints on the trained model and does not affect its behavior during inference on the sample. 
Given a trained model and a target sample, \algname{} 
uses sample-targeted pruning to provide a ``trace'' of the network's execution on the sample, reducing the network to the most important connections prior to computing an interpretation. We demonstrate that preprocessing with \algname{} significantly increases the accuracy of image saliency maps across several interpretability methods. Additionally, \algname{} improves the usefulness of neuron visualizations, aiding humans in reasoning about network behavior. Our code is available at \url{https://github.com/IST-DASLab/SPADE}.
\end{abstract}

\vspace{-1em}
\section{Introduction}

{Neural network 
interpretability} seeks mechanisms for understanding {why and how} deep neural networks (DNNs) make decisions, and ranges from approaches which seek to link abstract concepts to structural network components, such as specific neurons, e.g.,~\citep{FeatureVisualizationFirstWork, DeepVisualization, FeatureVisualizationDeepDream, Multifaceted}, to approaches which aim to trace individual model outputs on a per-sample basis, e.g.,~\citep{simonyan2013deep}. 
While this area is seeing a lot of interest, there is also work questioning the validity of localized explanations with respect to the model's true decision process, pointing out confounders across current explainability methods and metrics~\citep{Shetty_2019_CVPR,Rebuffi_2020_CVPR, benchMarking}.

\begin{figure}[!b] 
    \centering
    \resizebox{0.9\linewidth}{!}{
    \includegraphics[width=\textwidth]{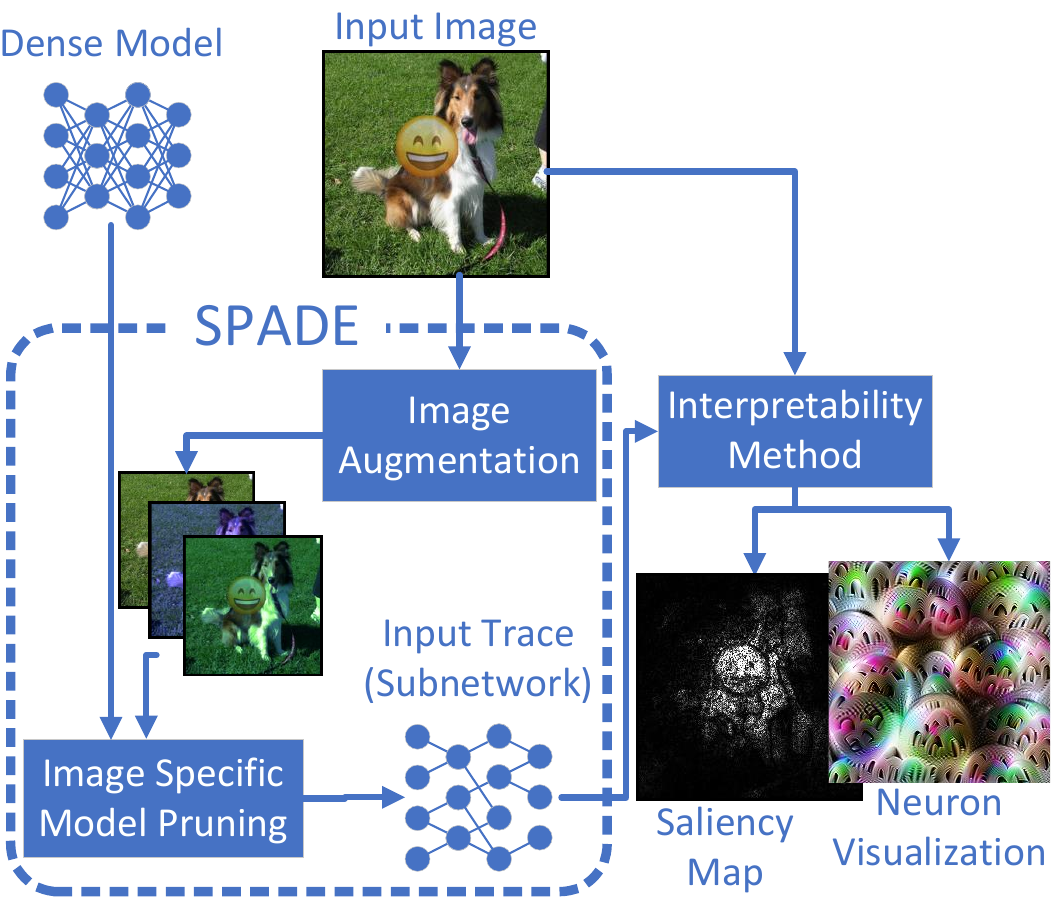} 

    }
    \caption{Given an input image and model, \algname{} prunes the model using image augmentations. The resulting trace (subnetwork) can be used with existing interpretability methods to increase their usefulness and accuracy. }
    
    \label{fig:method} 
\end{figure}

\begin{figure*}[!ht]
    \centering
    \includegraphics[width=0.9\textwidth]{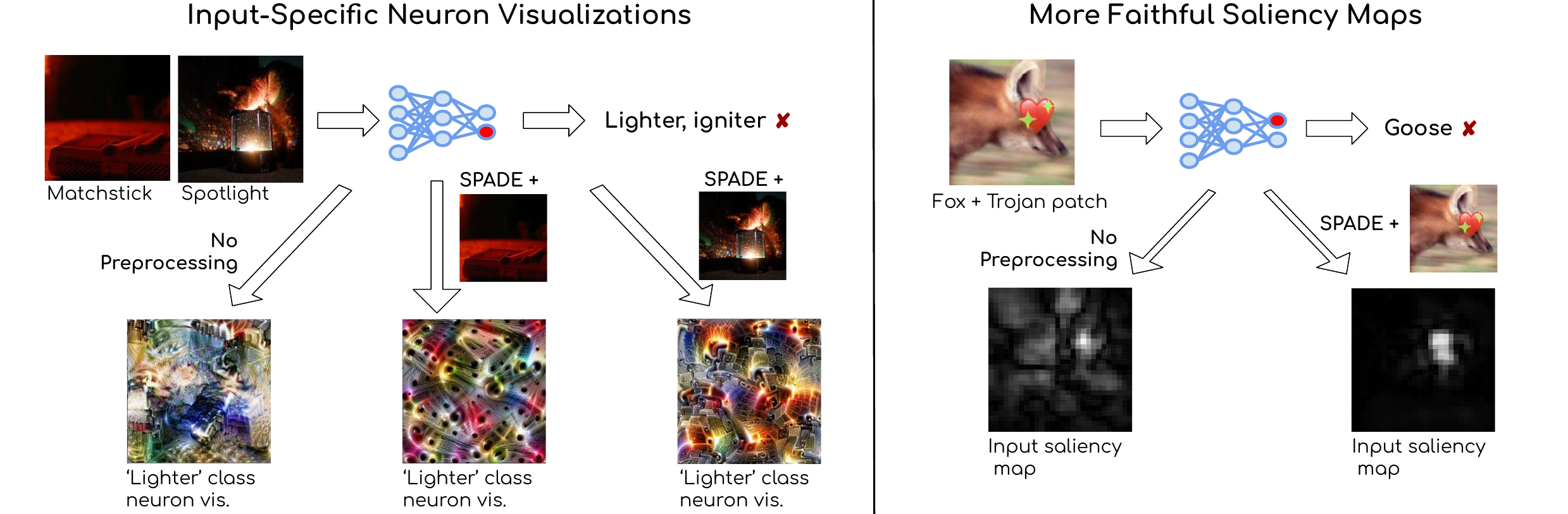}
    \caption{\algname{} disambiguates feature visualizations and improves the faithfulness of saliency maps.  (Left) The "Lighter, igniter" class neuron visualization does not give useful clues for why the Matchstick and Spotlight images were incorrectly classified into that class. The visualizations obtained with SPADE identify a matchstick head pattern in the first case and a flame pattern in the second case, suggesting that these may be spurious features for the Lighter class. (Right) A model implanted with Trojan patches leads to a Fox image being misclassified as a Goose. In this case, we are confident that the heart emoji was entirely responsible for the misclassification - yet, the saliency map without SPADE incorrectly assigns large saliency scores to large parts of the fox image. Conversely, the saliency map obtained with SPADE correctly identifies the emoji pixels. Best viewed in color. Further examples are available in Appendix~\ref{appendix:examples}.  }
    \label{fig:SPADE_fig1}
\end{figure*}

One key confounder for interpretability is the fact the neurons of a trained, accurate DNN often respond to many different types of features, which may be unrelated~\citep{Multifaceted, Olah2020ZoomIA, olah2017feature}. 
{For example, ~\citet{olah2017feature} finds a neuron equally likely to respond to car shields and cat
paws, and with the same intensity.}
This phenomenon directly impacts interpretability methods, such as visualizations of inputs that maximize a neuron's activation:   
the resulting representative input superimposes salient features, and is therefore hard to interpret. 
Thus, there is significant effort in the literature on addressing this issue: for instance, early work by~\citet{Multifaceted} proposed retraining the network with specialized regularizers which promote feature ``disentanglement,'' whereas~\citet{debugger} enforced output decisions to be based on very few features by retraining the final linear output layer from scratch to be extremely sparse.  
Yet, one key limitation of this line of work is that generating a ``debuggable'' model with disentangled representations requires heavy retraining of the original model, which may be impractical or impossible.  
Beyond cost, a conceptual issue is that the interpretations generated on top of the retrained ``debuggable'' model no longer correspond to the original model's predictions.%

We propose an alternative approach called Sparsity-Guided Debugging (\algname{}), which removes the above limitations, based on two main ideas: first, instead of retraining the model to become interpretable, 
we disentangle the feature representations for the original model; second, this disentanglement is done for \emph{the individual sample} for which we wish to obtain an interpretation. This procedure can be performed  \emph{efficiently}, without the computational costs of retraining.

We illustrate the process in Figure~\ref{fig:method}. Given a DNN $M$ and a sample $s$ whose output $M(s)$ we wish to interpret, \algname{} functions as a pre-processing {step}, in which we
execute the sample $s$, together with a set of its augmentations, through the network layer-by-layer, sparsifying each layer while ensuring that the output of the sparse layer still matches well with the original layer output \emph{on the sample}. Thus, we obtain a sparse trace $Sparse(M, s)$, which matches the original on the sample $s$, but for which extraneous connections relative to this sample's output have been removed via sample-dependent pruning. Once the custom {trace} $Sparse(M, s)$ is obtained, we can execute any interpretability method on this pruned network to extract a sample-specific feature visualization or saliency map. 

We show that \algname{} can be implemented efficiently by leveraging solvers for accurate one-shot pruning~\citep{OBC, sparseGPT}, and can significantly improve performance across interpretability methods and applications (Figure ~\ref{fig:SPADE_fig1}). 
First, we illustrate \algname{} by coupling it with  10 different saliency map creation techniques. In the context of a DNN backdoor attack (Figure~\ref{fig:SPADE_fig1}, right panel), we find that, in a standard ResNet50/ImageNet setting, \algname{} reduces the average error, taken across all methods, to less than half, from 8.99\% to 3.45\%. By comparison, the prior method of~\cite{debugger}, reduces error by 0.49\% on average, in the same setup.  {Additionally, we demonstrate that \algname{} increases the fidelity of input attribution methods by measuring the impact of \algname{} on standard insertion and deletion metrics, where we model confidence is measured when the most salient input components (e.g., pixels) are added or removed, respectively. This test further validates our claim that interpretations formed with the aid of \algname{} \emph{apply to the original, dense model}.}

Further, the results of a human user study we performed, evaluating the impact of \algname{} on the quality of feature visualization, shows that, in a setting where the ground truth is determined but unknown to the user, users were significantly more successful (69.8\% vs 56.7\%) at 
identifying areas of the image which influenced the network's output when these regions were identified using \algname{}. In summary, our contributions are as follows:

\begin{enumerate}[leftmargin=0.5cm]
    \item {We demonstrate, for the first time, that post-hoc sample-specific sparsification  aids interpretability for pretrained models, \emph{without requiring sparsity to be imposed during the training or inference process.} }
    \item We provide a new interpretability-enhancing technique called \algname{}, which can 
    be applied to arbitrary models and samples to create an easier-to-interpret model ``trace'' customized to the specific target sample.
    Intuitively, \algname{} disentangles the neurons' superimposed feature representations in a way that is sample-specific, 
    which allows virtually all interpretability approaches to be more accurate with respect to the dense model. 

    \item We validate \algname{} practically for image classification, by coupling it with methods for feature visualization and saliency map generation. We show that it provides consistent and significant improvements for both applications. Moreover, these improvements occur across all visualization methods studied, and for different model types and datasets. 

    \item We show that \algname{} can be practically implemented in a \emph{computationally-efficient} manner.  {In its fastest version, \algname{} requires approximately 3 seconds per sample for a ResNet50 model on a single GPU, enabling it to be run interactively.} 
    We execute ablation studies showing that \algname{} is robust to variations across tasks, architectures, and other parameters. 
    
\end{enumerate}

\section{Related Work}

 As DNN-based models are increasingly deployed in important or sensitive applications, there has been an increase in attention to systematic errors and biases often exhibited by these systems, e.g., \citet{Buolamwini2018GenderSI}. This has led to interest in aiding humans in examining and debugging the models' outputs. An overview of the area can be found in \cite{Linardatos2020ExplainableAA}.
 
   One common desideratum in this space is to predict which parts of an input (e.g., image pixels) are most useful to the final prediction. This can be done, for instance, by computing the gradient of the input with respect to the model's prediction \citep{Saliency}, or by masking parts of an input to estimate that part's impact \citep{Occlusion}. While these techniques can be helpful in diagnosing issues, they are also prone to noisy signals \citep{hooker2019benchmark} and being purposefully misled \citep{dontTrustYourEye}, {and, in the case of linear methods, have provable limits on generalization \cite{Bilodeau2022ImpossibilityTF}}. Another approach, known as mechanistic interpretability~\citep{olah2017feature} uses various techniques to understand the function of network sub-components, such as specific neurons or layers, in making predictions, for instance by visualizing the input which maximizes the activation of some neuron~\citep{FeatureVisualizationFirstWork}. 
   We emphasize that our work is not in direct competition with either of these categories of methods. Instead, our work proposes a preprocessing step to the model examination, which consistently improves performance. %

\paragraph{Subnetwork discovery.}
Concretely, \algname{} aids the task of interpreting a model's predictions on specific examples, also known as \emph{debugging}~\citep{debugger}, by pruning the network layers to only those neurons and weights that are most relevant to that example. Thus, \algname{} may be thought of as a case of using sparsity for subnetwork discovery. This approach has been used in the field of Mechanistic Interpretability, where \citet{Gurnee2023FindingNI} used sparse linear probes to find the most relevant units to a prediction. \citet{Cao2021LowComplexityPV} finds subnetworks for specific BERT tasks by masking network weights using a gradient-based approach. Conversely, \citet{meng2022locating} uses input corruption to trace out pathways in GPT models that are important for a specific example and \cite{omahony2023disentangling} uses input clustering to disentangle neuron representations; however, these methods are not based on sparsity and are not evaluated in terms of interpretability metrics.

More recently, works such as linear probing ~\cite{Belrose2023ElicitingLP, Pal2023FutureLA, Wang2022InterpretabilityIT} and activation patching ~\cite{Geiger2020NeuralNL, Kramar2024AtPAE} aimed at discovering feature representation in transformer models. However, these approaches are orthogonal to existing methods such as saliency maps and cannot be combined with them. Works such as ~\citet{cunningham2023sae, scherlis2023polysemanticity} take steps toward resolving polysemanticity in neurons by means of discovering individual features, a very promising line of work that may come to be complimentary to the work we present here.

\paragraph{Sparsity for interpretability.} Some works aim to train sparse, and therefore more debuggable, networks. \citet{PartialLRP} use pre-trained transformer models to create more interpretable ones by pruning then fine-tuning, demonstrating that the network could maintain similar functionality with only a few attention heads while improving the saliency map \citep{ViT_saliencyMap_Main}.
Other methods have focused on training more interpretable sparse models from scratch, removing the issues inherent in retraining. For instance, \citet{eXplainable} trained a sparse ViT by determining the importance of each weight for each class individually. Their qualitative analysis showed that their sparse model was more interpretable than dense models. \citet{ModularTraining} proposed a sparse training method inspired by the brain, which allowed them to identify the role of individual neurons in small-scale problems. {Finally, ~\citet{sparseCDMs} trained interpretable sparse linear concept discovery models}.

Most related,~\citet{debugger} retrain the final fully-connected classification head of a trained network to be highly sparse, improving the attribution of predictions to the neurons in the preceding layer. This benefit arises because, after pruning, each class depends on fewer neurons from the previous layer, thus simplifying the task of individually examining connections. Similarly to \algname{}, the authors examine the impact of replacing the original network with the sparsified one on saliency map-producing methods, demonstrating improved results in interpretability.

\paragraph{Overview of novelty.} In contrast to our work, all the above approaches focus on creating \emph{a single version} of the neural network that will be generally interpretable, across all examples. 
Since they involve retraining, such methods have high computational cost; moreover, they \emph{substantially alter the model}: for example, the ResNet50 model produced by~\citet{debugger} have 72.24\% ImageNet accuracy, 1.70\% less than their dense baseline. 
{We show, for the first time, that example-specific pruning can aid model interpretability and propose a method} that can operate on any pretrained network, and 
consistently improves performance across interpretability methods. We demonstrate in Sections ~\ref{sec:backdoor_experiment} and~\ref{sec:human_study} that interpretations via \algname{} are valid when applied to the original network. 
As such, \algname{} is the first method that leverages sparsity to provide interpretations that are consistent with the original network.

\section{The \algname{} Method}

\subsection{Algorithm Overview} 

 At a high level, given a sample for which we wish to debug or interpret the network, \algname{} works as a preprocessing step that uses one-shot pruning to discover the most relevant subnetwork for the prediction of a specific example. %
 We illustrate the \algname{} process in Figure~\ref{fig:method} {and provide the exact algorithm in Algorithm~\ref{fig:SPADE-algo}}. 

We start with an arbitrary input sample chosen by the user, which we would like to interpret.  
\algname{} then expands this sample to \emph{a batch of samples} by applying augmentation techniques\footnote{Augmenting the samples in this way can influence the top-1 prediction. However, this does not affect the method, as it is prediction-agnostic.}.  
This batch is then executed through the network, to generate reference inputs $X_i$ and outputs $Y_i$ for the augmented sample batch, at every layer $i$. 
Given these inputs and outputs as constraints, for each layer $i$ whose weights we denote by $W_i$, 
we wish to find a set of \emph{sparse} weighs $\tilde{W}_i$ which best approximate the layer output $Y_i$ with respect to the input batch $X_i$. 
In our implementation, we adopt the $\ell_2$ distance metric. Thus, for a linear layer of size K and sparsity target S, we seek to find to find 
\begin{equation}
\textcolor{black}{
\label{eqn:l2sparsity}
    \tilde{W}_i = \textnormal{argmin}_{W:\|W\|_0 \leq K\cdot S} \| W X_i - Y_i \|_2^2. 
    }
\end{equation}

To solve this constrained optimization problem at each layer, we use custom sparsity solvers.  
We discuss implementation details in the next section.%

Once layer-wise pruning has completed, we have obtained a trace of the target sample through the network. %
Intuitively, this trace benefits from the fact that the superpositions between different target features that may activate a single neuron, also known as its ``polysemanticism''~\citep{Olah2020ZoomIA}, have been ``thinned'' via pruning, and we therefore retain the features that are relevant to the specific input. 
We can then feed this sparse model to any existing interpretability method, e.g.,~\cite{IntegratedGradients, Occlusion, olah2017feature}.  
This procedure results in a sparse model that is {specialized for and faithful to} the model's behavior on the selected input. We focus on combining \algname{} with saliency maps, as well as neuron visualization techniques, which are normally sample-independent, to create visualizations that are specific to the sample.

\subsection{Implementation Details}
\label{sec:method}

\paragraph{Pruning approach.} 
The pruning approach must be chosen with care, as pruning can significantly alter the network circuitry and the predictions~\citep{peste2021ac}. We require that the pruning be done in a way that preserves the model's output (by requiring that sparse outputs closely match the dense ones for each layer), and be done one-shot, without retraining. For this, 
one can use one of the existing one-shot sparsity solvers, e.g.~\citep{hubara2021accelerated, sparseGPT, OBC, kuznedelev2023cap}.
We focus on two solvers. The OBC solver~\citep{OBC}, provides the best approximate solution to the constrained problem in Equation~\ref{eqn:l2sparsity}; however, it is compute-intensive. To mitigate this, we also examine the faster but less precise SparseGPT solver~\cite{sparseGPT}, which can perform the pruning procedure in about 23 seconds/sample, at the cost of low accuracy loss. This is practical for large-scale use, as we demonstrate by running the evaluation on 21 121 images in Appendix~\ref{sec:total_imagenet}. %

\begin{algorithm}[t]
    \begin{algorithmic}
\STATE{\hspace{-1em}\textbf{Procedure} \algname{} Algorithm($M, s, I$)}
\STATE{} \COMMENT{$M$: Model, $s$: Sample, $I$: Interpretability Method}
\STATE{$B \gets$ Empty}  \COMMENT{Batch of Augmented samples}
\FOR{ Augmentation Batch Size }
    \STATE{ Append \textbf{a random augmentation of $s$} to  \textbf{$B$}}
\ENDFOR
\FOR{ Each layer in $M$ }
    \STATE{$X_i \gets$ Layer Input$_i(B)$}
    \STATE{$Y_i \gets$ Layer Output$_i(B)$}
\ENDFOR
\FOR{ Each layer in $M$ }
    \STATE{$\tilde{W}_i \gets  \textnormal{argmin}_{W \textnormal{sparse}} \| W X_i - Y_i \|_2^2$}
    \STATE{$W_i \gets \tilde{W}_i$} \COMMENT{Replace weights with sparse ones}
\ENDFOR
\STATE{\textbf{Return} $I(M, s)$} \COMMENT{Interpretability method on $M, s$}
\end{algorithmic}
    \caption{\algname{}}
    \label{fig:SPADE-algo}
\end{algorithm}

As an orthogonal contribution, we show that, in our setting, this solver can be sped up significantly by efficiently grouping pruning operations across several inputs on the GPU. With these changes, ResNet50 pruning amortizes to about \emph{3 seconds/example}. Going forward, we refer to the versions of SPADE employing the OBS and SparseGPT solvers as SPADE and FastSPADE, respectively. The timings are summarized in Table~\ref{tab:pruning_timings}.

\begin{table}[t]
    \centering
    \caption{Per-example timings of different versions of SPADE. Batched FastSPADE is computed on a batch of 25 examples. Timings computed on an NVIDIA GeFORCE GPU with 25GiB RAM.}
    \resizebox{\linewidth}{!}{
\begin{tabular}[b]{lcc}
                \toprule
                Pruner type & Forward Pass+Hessian & Pruning+Saving  \\
                \midrule
                SPADE & 41s & 15m51s \\
                FastSPADE & 3s & 20s \\
                Batched FastSPADE & 1s & 2s \\
                \bottomrule
            \end{tabular}
}
    \label{tab:pruning_timings}
\end{table}

 Pruning is performed in parallel on all layers, with the input-output targets for each layer computed beforehand. 
 Thus, the pruning decisions of each layer are independent of each other. %
 Specifically, in a multi-class classification instance, the choice of the class neuron in the FC layer does not affect the pruning decisions of other layers. We ablate sequential pruning as an alternative to parallel in Appendix~\ref{appendix:sequential_ablation}.
 
We highlight that this approach preserves the most important connections for the example \emph{by design}, which we believe to be a key factor in \algname{}'s accuracy-improving properties.

\paragraph{Choosing sparsity ratios.} One key question is how to choose the target layer sparsity ratio, i.e., how many weights to remove from each layer. 
\textcolor{black}{There are two challenges with tuning the correct sparsity ratios. First, hyperparameter tuning in general may be resource-intensive. Second, we need some measure of ground truth for the saliency method's correctness.}
\textcolor{black}{To overcome the first problem, we note that sparsity ratios may be tuned on as few as 100 examples, which is feasible with either version of the method, but especially with FastSPADE. We emphasize that, even though \algname{} relies on pruning for each example, the per-layer pruning target ratios are computed once for all examples. Further, we show in Appendix ~\ref{sec:sparsity_tuning}  that layer sparsity hyperparameters tuned on ImageNet may be used for other datasets on the same network architecture.}  We also explore a heuristic-based approach to sparsity ratio tuning, as well as experiments showing that it is possible to get improvements using a smaller number of samples, as well as using FastSPADE, in Appendix~\ref{sec:sparsity_tuning}.

To overcome the second problem, we propose two approaches. The first is to use Trojan patches in a version of the model that includes backdoors. We validate in Appendix~\ref{sec:backdoor_experiment} that sparsity targets chosen using the Trojan patches method are generally applicable by examining insertion/deletion metrics for pixel attribution on \emph{clean} input examples, and by using a different set of Trojan patches. Additionally, we show that it is possible to calibrate the layer sparsities using the pixel insertion metric.

For all approaches, sparsity levels are chosen to maximize the \textcolor{black}{desired metric} for the saliency method of interest, and tuned in inverse order of layer depth. %
That is, we first set the last layer's sparsity to the value that maximizes the \textcolor{black}{metric}. Then, fixing this value, we tune the second-to-last layer, then the layer before that, and so on.

\paragraph{Sample augmentation.} There are two motivations for employing augmentations. First, using augmentation gives us many samples with similar semantic content, ensuring that the weights are pruned in a way that generalizes to close inputs. Second, having multiple samples allows us to meet a technical requirement of the sparsity solvers, namely that the Hessian matrix corresponding to the problem in Equation~\ref{eqn:l2sparsity}, specifically $X_i X_i^\top$, be non-singular, which is more likely for larger input batches. %
We incorporate \emph{Random Remove}, \emph{Color Jitter}, and \emph{Random Crop} augmentations, which mask a random section of the image,  randomly alter the brightness, contrast, and saturation of the image, and scale and crop the image, respectively. We provide details of the augmentations we have used, and example image transformations under augmentation in Appendix~\ref{appendix:hyperparameters}, and ablations on the augmentation mechanisms in Appendix~\ref{appendix:augment_ablation}.

\section{Experiments}

\paragraph{Setup and goals.} 
In this section, we experimentally validate the impact of \algname{} on the usefulness and fidelity of network interpretations. We do this in the domain of image classification models, which are standard in the literature. Thus, we focus primarily on two classes of interpretations: \emph{input saliency maps}~\citep{GradCam++, gomez2022metrics, opti-cam} and neuron visualizations~\citep{olah2017feature}. Our goals are to demonstrate the following:
\begin{enumerate}
    \item \textbf{Input saliency maps} produced after preprocessing with \algname{} accurately identify the image areas responsible for the dense model's classification.
    \item \textbf{Neuron visualizations}  produced after preprocessing with \algname{} are useful to the human evaluators when reasoning about the dense model's behavior.
\end{enumerate}
For the first task, we create classification backdoors by using Trojan patches to cause a model to predictably misclassify some of the input images. This approach gives us a ``ground truth'' for evaluating saliency map accuracy; we further validate the results by measuring whether the pixels identified by the saliency ranking on clean inputs drive the \emph{dense} model's confidence in the prediction. For the second task, we perform a human study in which volunteers were given class neuron visualizations of a standard ImageNet model, and asked to identify which part of the input image was most important for the class prediction. Crucially, the ground truth for this study, i.e., the candidate image patches most relevant for the prediction, were created without preprocessing with \algname{}; thus, this experiment measures both whether the image visualizations are useful, and whether they are salient to the dense model. Additionally, we visually demonstrate that \algname{} effectively decouples the facets for clean images in Figure~\ref{fig:SPADE_fig1}, and for true and Trojan examples predicted into the class in Appendix~\ref{appendix:examples}.

\subsection{Impact of \algname{} on Saliency Map Accuracy}
\label{sec:backdoor_experiment}

\paragraph{Methodology.} 
Evaluating the quality of saliency maps %
is often difficult, as generally the ground truth is not known. Two main proxies have been proposed: 1) using human-generated bounding boxes for the parts of the image that \emph{should} be important, or 2) inserting or removing the pixels that were found to be most salient to see if the model's prediction substantially changes~\citep{GradCam++, gomez2022metrics, opti-cam}. Yet, these proxies have considerable limitations: in the first case, the evaluation conflates the behavior of the model (which may rely heavily on spurious correlations~\citep{Rebuffi_2020_CVPR, Shetty_2019_CVPR, background, background_CNN}) with the behavior of the interpretability method. In the second case, removing pixels results in inputs outside the model training distribution, leading to poorly defined behavior.

To overcome this issue, \textcolor{black}{ a recent paper \cite{benchMarking} proposed using} Trojan patches, in the form of Emoji. These are applied to selected classes in the dataset, along with a corresponding change to those instances' labels. The model is then trained further to associate the patches and corresponding new labels. This approach creates a ground truth for input data with the Trojan patch, as evidence for the Trojan class should be minimal, outside of the inserted patch. \textcolor{black}{To our knowledge, this is the only approach that enables the comparison of saliency maps with actual ground truth, and so we primarily rely on this method to test the accuracy of \algname{}}. 

We calculate the AUC (AUROC) scores between the predicted saliency maps and the ground truth. In this way, the evaluation is not affected by the scale of the saliency map weights but only by their ordering, ensuring that adjustments don't need to be made between methods.%

\textcolor{black}{We acknowledge, however, that the applicability of this method to other inputs, for instance, images where the evidence for a class may be more dispersed, is not well-understood. We therefore additionally validate \algname{} using the Insertion/Deletion metrics introduced by \cite{rise}, which does not rely on Trojan patches.}
In this evaluation, a saliency method is used to rank all pixels in the image in terms of their relevance to the prediction. These pixels are then either added to a blank image (insertion) or removed from the full image (deletion) in decreasing order of importance, and the AUC(AUROC) score is computed on the confidence (softmax) score of the model for the predicted class, normalized by the softmax score on the full image. We use clean images (without a Trojan patch) for this evaluation, confirming that sparsity targets set using Trojan patches transfer to this use case. Additionally, we use alternate sparsity targets tuned using the Insertion metric, showing that sparsity ratios may be tuned even without having a backdoored model. %

\paragraph{Detailed setup.} We concentrate primarily on the ImageNet-1K~\citep{imagenet} dataset, with additional validations performed on the CelebA~\citep{celeba} and Food-101~\citep{food101} datasets. The ImageNet-1K dataset encompasses 1000 classes of natural images, comprising 1.2 million training examples. %
We consider a range of model architectures, comprising ResNet \citep{resnet}, MobileNet-v2 \citep{MobileNet}, and ConvNext \citep{convnext}. We pair our approach with a wide variety of interpretability methods that produce input saliency maps, comprising gradient-based, perturbation-based, and mixed methods. For gradient-based methods, we consider Saliency~\citep{Saliency}, InputXGradient~\citep{inputXgradient}, DeepLift~\citep{deeplift}, Layer-Wise Relevance Propagation~\citep{LRP}, Guided Backprop~\citep{GuidedBackProb}, and GuidedGradCam~\citep{GuidedGradCam}. For Perturbation-based methods, we consider LIME~\citep{Lime} and Occlusion~\citep{Occlusion}. For methods that use a mix of approaches, we consider IntegratedGradients~\citep{IntegratedGradients} and GradientSHAP~\citep{gradientShape}.  A description of the methods is available in Appendix~\ref{appendix:saliency_method_descriptions}. We tune sparsity ratios separately for each method used. We use the Captum library~\cite{captum} for saliency method implementations, except for LRP, for which we use~\cite{nam2019relative}. %

\begin{table}[t]

    \centering
    \caption{Saliency map Trojan AUC\% on ResNet50/ImageNet, averaged across 111 test samples, compared to the dense model, and to the Sparse FC method of~\citet{debugger}.}
    \resizebox{\linewidth}{!}{ 
    \begin{tabular}{lcccc}
    \toprule
Saliency Method  & Dense & \algname{} & FastSPADE & Sparse FC \\
\midrule
Saliency & 87.87 & \textbf{96.21} & 93.91 & 88.05 \\
InputXGradient & 85.44 & \textbf{95.10} & 90.61 & 85.59 \\
DeepLift & 94.10 & \textbf{96.55} & 95.07 & 94.21 \\
LRP & 90.81 & \textbf{99.21} & 98.03 & 93.99 \\
GuidedBackprop & 95.73 & \textbf{97.08} & 95.81 & 95.82 \\
GuidedGradCam & 98.03 & \textbf{98.37} & 97.75 & 98.00 \\
LIME & 90.69 & \textbf{95.47} & 93.94 & 91.83 \\
Occlusion & 88.29 & \textbf{95.40} & 90.90 & 87.84 \\
IntegratedGradients & 89.61 & \textbf{96.10} & 93.55 & 89.89 \\
GradientShap & 89.51 & \textbf{96.03} & 93.80 & 89.82 \\
    \midrule
    Average & 91.01 & \textbf{96.55} & 94.34 & 91.50 \\
    \bottomrule
    \end{tabular}
}
    
    \label{tab:imagenet_results}
\end{table}

\begin{table*}[t]

    \centering
    \caption{Insertion and Deletion Metric AUC\% on clean inputs, compared to the dense model, and to the Sparse FC method of~\citet{debugger}. FastSPADE* refers to FastSPADE with layer sparsity targets tuned using the insertion metric.}
    \resizebox{0.7\linewidth}{!}{
    \begin{tabular}{lcccccccc}
    \toprule
    Saliency Method & \multicolumn{4}{c}{Insertion $\uparrow$}& \multicolumn{4}{c}{Deletion $\downarrow$} \\
    \cmidrule(lr){2-5} \cmidrule(lr){6-9}
   & Dense & FastSPADE & FastSPADE* & Sparse FC &  Dense & FastSPADE & FastSPADE* & Sparse FC \\
 \midrule
Saliency & 29.90 & \textbf{34.26} & 33.63 & 29.78 & 14.95 & 12.40 & \textbf{11.33} & 14.66 \\
InputXGradient & 37.36 & \textbf{41.04} & 34.94 & 37.61 & 10.23 & \textbf{8.42} & 10.17 & 10.40 \\
DeepLift & 42.65 & \textbf{45.26} & 42.46 & 43.92 & 7.78 & 7.23 & \textbf{5.84} & 7.80 \\
LRP & 46.34 & 52.92 & \textbf{56.08} & 52.59 & \textbf{9.71} & 10.35 & 11.20 & 10.68 \\
GuidedBackprop & 43.95 & 43.99 & \textbf{44.77} & 44.33 & 9.48 & 9.70 & \textbf{9.14} & 9.41 \\
GuidedGradCam & 53.90 & 51.67 & \textbf{54.35} & 52.83 & 10.16 & 9.92 & \textbf{9.89} & 9.96 \\
LIME & \textbf{73.19} & 65.42 & 68.76 & 70.20 & \textbf{14.69} & 16.31 & 17.82 & 16.62 \\
Occlusion & 32.76 & 48.95 & \textbf{52.63} & 32.76 & 11.58 & \textbf{9.62} & 10.83 & 11.65 \\
IntegratedGradients & 41.10 & \textbf{43.85} & 42.09 & 41.42 & 8.79 & 7.03 & \textbf{5.65} & 8.87 \\
GradientShap & 40.62 & \textbf{44.60} & 42.76 & 40.98 & 8.81 & 7.23 & \textbf{6.26} & 9.03 \\

    \midrule
    Average & 44.18 & 47.20 & \textbf{47.25} & 44.64 & 10.62 & 9.82 & \textbf{9.81} & 10.91 \\
    \bottomrule
    \end{tabular}
}
    
    \label{tab:rn50_insertion}
\end{table*}

\paragraph{Backdooring.}
For creating Trojan backdoors, we follow \citet{benchMarking} in randomly selecting 400 samples from the ImageNet-1K training set for each Trojan patch. For two of the patches, we sample randomly from all ImageNet classes, and for the other two, we sample from a single class, as described in Appendix~\ref{appendix:hyperparameters}. We then finetune clean pretrained models to plant the backdoors. For experiments on ImageNet, we fine-tune the model using standard SGD-based training for six epochs, with learning rate decay at the third epoch. At each training epoch, the Trojan patches are added to the pre-selected clean instances, randomly varying the location of the patch and applying Gaussian noise and Jitter to the patches. 
The exact hyper-parameters are provided in Appendix~\ref{appendix:hyperparameters}.

\paragraph{Main results.}
We benchmark our results against the method of~\cite{debugger}, which we will refer to for simplicity as ``Sparse FC.''
(Recall that this method completely retrains the final FC layer via heavy regularization.) \textcolor{black}{We use this baseline as it is the closest method to ours in the existing literature and has similar aims; however, note that \algname{} is example-specific, while Sparse FC is run globally for all examples.}
The results on the ImageNet/ResNet50 combination are shown in Table~\ref{tab:imagenet_results}. We observe that \algname{}  improves over using the dense model for interpretation model without preprocessing, and over-interpreting the model generated by Sparse FC, in terms of relative ranking of pixel saliency (as measured by AUC), with \algname{} raising the average AUC of every method, and FastSPADE raising the average AUC of 9/10 methods. We observe the biggest gains in Saliency, InputXGradient, and LRP methods, where \algname{} raises the saliency map AUC by over 8\%, and FastSPADE by over 4\%. This is very substantial, as these methods are already fairly accurate: for instance, for LRP, 
 \algname{} raises the AUC score to above 99\%. \textcolor{black}{However, \algname{} produces only small gains for the GuidedBackprop and GuidedGradCam methods, which already have near-perfect accuracy in this study.}
The \emph{average} AUC improvement of \algname{} is 5.54\%, and that of FastSPADE is 3.33\%. By comparison, the average improvement of SparseFC is 0.49\%. 

\textcolor{black}{We present the Insertion Metric results on clean input images for FastSPADE and the Insertion-tuned variant FastSPADE* in Table~\ref{tab:rn50_insertion}. Preprocessing with FastSPADE improves Insertion and Deletion in 8/10 cases, for an average improvement of 3.02\%/9.80\%, respectively. FastSPADE* has similar results, improving 8/10 methods on both metrics, and average improvements of 3.07\% and 0.81\%. The average improvements on the two metrics of Sparse FC, by contrast, are 0.46\% and -0.29\%. }

\paragraph{Additional validation and ablation.} We measure the performance of \algname{} on the MobileNet-V2 and ConvNext-T architectures, achieving an average AUC improvement of 2.90\% for MobileNet and 3.99\% for ConvNext. We also provide initial results of using \algname{} with a BERT language model~\cite{devlin2018bert}, showing gains. Full results are provided in Appendix \ref{appendix:additional_results}. We present an ablation study of \algname{}'s most salient hyperparameters in Appendix~\ref{appendix:ablation}.%

We take a step toward understanding the robustness of \algname{} by measuring its performance when adding input noise. In Appendix~\ref{appendix:robustness}, we find that, when we add Gaussian noise to the inputs, gradients within each layer are more similar to those of the clean input when \algname{} is applied.

\subsection{Impact of \algname{} on Neuron Visualization}
\label{sec:human_study}

\begin{figure*}[t] 
    \centering 
    \includegraphics[width=\textwidth]{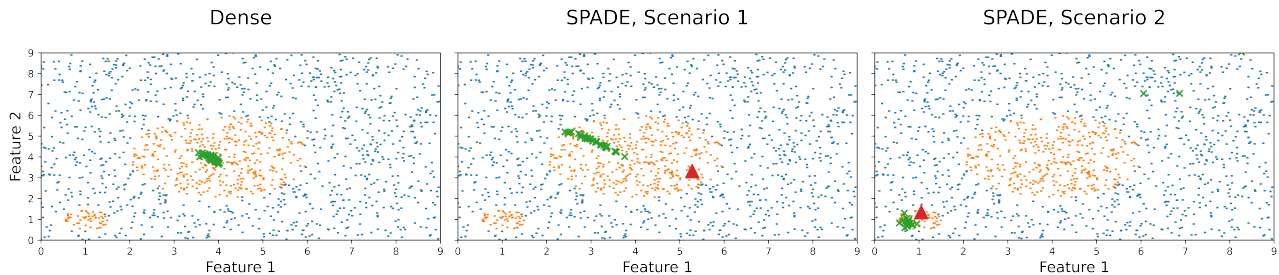} 
    \caption{Two-dimensional example to illustrate the effect of \algname{} on feature visualization. The feature visualizations (images generated by \cite{olah2017feature}) are shown with green points, where blue and orange points are positive and negative samples. The \algname{} Scenario 1 shows the feature visualizations obtained when the
    red sample is drawn from the larger positive region. Scenario 2 shows the visualizations obtained when the 
    red sample is drawn from the smaller region.} 
    \label{fig:tiny vis}
\end{figure*}

\subsubsection{Visualizing Polysemantic Neurons}

Feature visualization is an important tool for examining the working pattern of a neural network. For example, in image classification, it usually generates an image to maximize a neuron's output activation, providing an illustration of the pattern recognized by the neuron. Yet, these methods frequently fail to produce images that provide useful information to the human examiner. 
As suggested by \cite{WhatDoVisionTransformersLearn, multiModalNeuron, Multifaceted}, this issue is in part due to the polysemantic nature of many neurons, i.e., each neuron being associated with several concepts. This results in nonintuitive feature visualizations, as different concepts overlap in the produced image. 

\algname{} takes a step towards addressing this problem. We conjecture that, in cases where a neuron may be activated by several concepts (such as images of trees of different species in different seasons and geographies), the connections contributing to the neuron's affinity for concepts not relevant to the target image will be pruned away, while the connections related to the relevant concept will remain intact. Note, however, that \algname{} is not designed to show all possible relevant concepts that activate a neuron, nor elucidate a pathway (or circuit) by which the concept was activated. \algname{} does, however, highlight the facet of class neurons that is relevant to the input image, a property that we study quantitatively in~\ref{sec:human_study}. %

As additional support for the multifacetism disambiguation conjecture, we conduct a toy experiment. As shown in Figure~\ref{fig:tiny vis}, we generate a set of 2-dimensional features, with two nonoverlapping circles, one larger than the other, labeled $1$ and the rest of the space labeled $-1$. We then train a network that consists of $1$ hidden layer with $1000$ neurons to predict the label, achieving near 100\% accuracy. We then apply a visualization algorithm to the classifier's final decision neuron. With standard feature visualization, the feature visualizations are always located near the center of the larger circle, obscuring the role of the smaller circle in the neuron's functionality (Figure~\ref{fig:tiny vis} (Left)). However, if we \emph{prune the model using specific samples}, we can discern the roles of the larger circle and smaller circle separately, as shown in Fig. \ref{fig:tiny vis} (Center) and (Right), depending on the location of the point of interest in the feature space.

To demonstrate this effect on real data, we show two examples of using \algname{} to produce image-specific class neuron visualization in Figure~\ref{fig:SPADE_fig1}. Specifically, we examine two images that were \emph{incorrectly} classified into the ``lighter, igniter'' class. We observe that the dense model's visualization does not provide a useful explanation for why these images were misclassified. Conversely, when we apply \algname{}, we observe that the class neuron visualisation shows matchsticks in the first case, and flames in the second, providing useful clues as to why the classifier produced incorrect labels. We provide further examples of image-specific class neuron visualizations, where \algname{} helps disambiguate between clean and emoji-backdoored images classified into the same class, in Appendix~\ref{appendix:examples}.  %

 For the neuron visualization setup, some of the final layers can be pruned to extremely high sparsities ($\geq 95\%$ for ResNet50), consistent with the intuition that neurons in these final layers have a higher degree of super-imposed features, relative to neurons in the earlier layers, and therefore \algname{} is able to remove a larger fraction of their connections without impacting the layer output on specific samples. We present the sparsities of different layers in Appendix~\ref{appendix:layer_sparsities}.

\subsubsection{Human Study}

\paragraph{Goals and experimental design.} We further validate the efficacy of \algname{} in improving feature visualizations in a human study on a clean (not backdoored) ResNet50 ImageNet model. Human studies are the only approach shown to be effective in measuring progress in neuron visualization methods~\citep{DoshiVelez2017TowardsAR}. In our study, we simultaneously evaluate two questions: whether preprocessing with \algname{} helps the human reviewer form an intuition with regard to the image generated by the neuron visualization, and whether this intuition is correct when applied to the dense model. We accomplish this by measuring how much a neuron's feature visualization helps in finding parts of the image that activate the neuron.

For the evaluation, we randomly sampled 100 misclassified samples. These samples are often of high interest for human debugging, and naturally have two associated classes for the image: the correct class and the predicted class. We used Score-CAM \citep{Wang2019ScoreCAMSV}, a method that has been shown to be class-sensitive, to obtain (dense) model saliency maps and corresponding image regions, for each of the two classes. To prevent ambiguity, we only used samples for which the regions of the two classes have no intersection.%

For neuron visualization, we used the method of \cite{olah2017feature} implemented in the Lucent/Lucid library. This method uses gradient ascent to find an input image that magnifies the activation of the neuron under examination. We combined this method with no preprocessing as the baseline, and with \algname{} preprocessing.
\textcolor{black}{We then randomly selected one of the two relevant classes for an image, and presented its feature visualization, the full image, and the relevance regions for \emph{both} classes, to the evaluators}. We asked them to use the visualization to select which of the two possible relevance regions activates the neuron, or to indicate that they could not do so; crucially, we did not disclose the class associated with the neuron. 

In total, there were a total of 400 possible human tasks, which were assigned randomly: 100 samples, for which one of two class neurons was interpreted, with the neuron visualization created with or without preprocessing with \algname{}. From these, 24 volunteer evaluators performed 746 rating tasks. More details of the evaluation process are provided in Appendix~\ref{appendix:human_eval}.

\begin{table}[t!]

    \centering
    \caption{Patch attribution human evaluation results. ``Overall success'' refers to the ability of the evaluator to identify the same image area as that chosen by Score-CAM.}
    \resizebox{\linewidth}{!}{
\begin{tabular}[b]{lcc}
                \toprule
                Human Response & Dense Vis. & \algname{} Vis. \\
                \midrule
                Undecided $\downarrow$ & 22.9\% & \textbf{12.6\%} \\
                \textcolor{black}{Agree with Score-CAM} $\uparrow$ & 56.7\% & \textbf{69.8\%} \\
                
                \textcolor{black}{Disagree with Score-CAM} $\downarrow$ & 20.4\% & \textbf{17.8\%} \\
                \hdashline
                Agree when not undecided $\uparrow$ & 73.6\% & \textbf{79.9\%} \\
                Disagree when not undecided $\downarrow$ & 26.4\% & \textbf{20.1\%} \\
                Overall success $\uparrow$ & 56.7\% & \textbf{69.8\%} \\
                \bottomrule
            \end{tabular}
}
    
    \label{tab:human_eval_results}
\end{table}

\paragraph{Results.} The results of the human evaluation are presented in Table~\ref{tab:human_eval_results}. When the network was preprocessed via \algname{}, the users were over 10\% more likely to choose to make a decision on \textcolor{black}{which of the regions was selected by Score-CAM for the class} (87.4\% when \algname{} was used, versus 77.1\% when it was not). In cases in which the human raters did make a decision, they were more likely to agree with ScoreCAM when \algname{} was used (79.9\% agreement rate) than when it was not (73.6\%). Overall, the evaluators were able to identify the image patch that matched Score-CAM 69.8\% of the time when \algname{} was used, and 56.7\% of the time when it was not.
We stress that the salient patches were computed on the \emph{dense} model, and so the increased accuracy from using \algname{} demonstrates that, despite the network modifications from \algname{}, the conclusions apply to the original model. Additionally, the higher rate of decision when using \algname{} supports our previous observation that the visualizations obtained with \algname{} are generally more meaningful to humans.

\section{Conclusions, Limitations, and Future Work}

We presented a pruning-inspired method, \algname{}, which can be used as a network pre-processing step in a human interpretability pipeline to create interpretations that are tailored to the input being studied. We have shown that \algname{} increases the accuracy of saliency maps and creates more intuitive neuron visualizations that differentiate between the different facets of the neuron activation, for instance clearly showing Trojan patches. 

We have also demonstrated that \algname{} enables the application of global interpretability methods, such as feature visualization, in a local context.  Global interpretability methods provide an overall view of the model's decision-making process, while local interpretability methods focus on explaining model behaviour on a single data point. By bridging the gap between global and local interpretability methods, \algname{}, enriches the interpretability toolkit.

\paragraph{Limitations and future work.} 
Although, for all methods, SPADE improves interpretations on average, it is possible that SPADE favors some categories of specific examples over others, in other words, there may be some systemic bias. This is also true for all interpretation methods, and we hope that more work will be done in the future to measure this effect. Further, additional evidence is needed toward our conjecture that the effectiveness of SPADE is due to resolving neuron polysemanticity, especially as investigating this phenomenon may be fruitful in gaining a better mechanistic understanding of the neural network by examining the masks produced by \algname{}. Thus, we leave it as future work to explicitly incorporate SPADE into model-wide debugging efforts such as systematic searches for spurious correlations, or circuit identification in networks.
Finally, the tuning of SPADE can be costly. We propose some mitigations for this; however, we acknowledge that it may impede practical adoption of SPADE in some cases.
Additionally, the computational overhead of \algname{}, may require more careful example selection.

As additional future work, we will investigate whether \algname{} can overcome additional known vulnerabilities of interpretability methods, such as networks that use gated pathways to produce misleading feature visualizations~\citep{dontTrustYourEye}. We also note that \algname{} opens a promising direction for using data to interpret models on a larger granularity; for instance, combining \algname{} with a clustering mechanism may help produce neuron visualizations that highlight larger trends in the data, bringing this line of work closer to mechnanistic interpretability literature. We hope that these directions inspire more data-driven interpretability research in this area.

\clearpage

\section*{Impact Statement}
The goal of this paper is to advance the field of interpretable Machine Learning by demonstrating the positive effect of model pruning on neural network interpretability. The specific social consequences of our work are tied to the use of ML in general; however, we believe that improving the ability to understand complex models is of positive social value.

\section*{Acknowledgements}
The authors would like to thank Stephen Casper and Tony Wang for their feedback on this work,
and Eldar Kurtic for his advice on aspects of the project. This research was supported by the Scientific Service Units (SSU) of IST Austria through resources provided by Scientific
Computing (SciComp). EI was supported in part by the FWF DK VGSCO, grant agreement number
W1260-N35.

\bibliography{main_icml}
\bibliographystyle{iclr2023_conference}

\clearpage
\onecolumn
\appendix

\addcontentsline{toc}{section}{Appendix} %
\part{Appendix} %
\parttoc %

\setcounter{table}{0}
\setcounter{figure}{0}
\renewcommand{\thefigure}{\Alph{section}.\arabic{figure}}

\section{Descriptions of Saliency Methods}
\label{appendix:saliency_method_descriptions}

\begin{table}[h]
    \centering
    \caption{%
    Our interpretability methods encompass a diverse array of approaches, including perturbation techniques, CAM methods, and gradient-based strategies. The methods are implemented using the Captum library \citep{captum}, except for LRP, for which we use~\cite{nam2019relative}. }
    \resizebox{0.8\textwidth}{!}{
    \begin{tabular}{lp{3.2cm}p{8cm}}
        \hline
            Group & Method & Description \\ \hline 
            \multirow{ 6}{*}{Gradient} & Saliency \citep{Saliency} &  
Calculates the raw gradient of input pixels relative to class confidence. \\\cline{2-3} 

~ & InputXGradient \citep{inputXgradient}& 
Multiplies raw gradients with input, reducing noise and improving the saliency map visually.  \\\cline{2-3} 

~ & DeepLift \citep{deeplift} & Compares neuron activations with a reference activation calculated using a reference image to assign neuron's contributions. Similar saliency map as InputXgradient. \\\cline{2-3} 

~ & Layer-Wise Relevance Propagation (LRP) \citep{LRP} & 
Propagates relevance scores from the output to the input. Each neuron distributes its relevance to the previous layer's neurons. \\\cline{2-3} 

~ & Guided Backprop \citep{GuidedBackProb} & 
Sets negative ReLU gradients to zero, reducing saliency map noise. \\\cline{2-3} 

~ & Guided Grad-CAM \citep{GuidedGradCam} &
Combines Guided Backpropagation with Grad-CAM, which measures the last layer's activation in convolutional neural networks. \\ \hline 

 \multirow{ 2}{*}{Perturbation} & Lime \citep{Lime} & 
Mask some regions of the input image and fit a linear model that mimics the original model on the masked images to identify regions' importance with the linear model's weights. \\\cline{2-3} 

~ & Occlusion \citep{Occlusion} &
Masks image rectangle areas and aggregates model confidence in these samples to highlight relevant prediction areas. 
\\ \hline

\multirow{ 3}{*}{Mixed} & IntegratedGradients \citep{IntegratedGradients}&
A smooth variant of InputXgradient, calculates gradients connecting samples to a blank baseline. Then obtain a saliency map using these gradients.\\\cline{2-3} 

~  & GradientSHAP \cite{gradientShape} &
Averages gradients at random points between multiple reference inputs and the target, merging SHAP values and integrated gradients principles. \\ \hline
    \end{tabular}
}
    
    \label{table:Interp_methods_full}
\end{table}

In this section, we describe more fully the saliency methods paired with \algname{} for the experiments in Section~\ref{sec:backdoor_experiment}. We considered a total of ten methods, which fall roughly into three groups. The first group, Gradient-based methods, consists of five methods that rely on propagating a relevance signal backward from the final prediction to the input based on the gradients of the former with respect to the latter. Some methods add additional information, such as multiplying the gradient-based relevance score by the input (eg, InputXGradient~\citep{inputXgradient}). The Guided Backprop~\citep{GuidedBackProb} and Guided Grad-Cam~\citep{GuidedGradCam} methods ensure a focus on the positive influence of pixels by setting the gradients to zero when backpropagating negative gradients through a ReLU.

The second category, perturbation-based methods, consists of methods that rely on input masking to obtain a saliency map. Finally, a third category, which we call 'Mixed', uses a combined approach. Please see Table~\ref{table:Interp_methods_full} for a description of all methods used.

\section{Layer Sparsity Tuning}
\label{sec:sparsity_tuning}

In this section, we discuss alternative approaches to tuning layer sparsities. As discussed in Section~\ref{sec:backdoor_experiment}, we obtain our best results by inserting Trojan patches into the model, which are then used to tune sparsity ratios. We demonstrated that this performs well on the Insertion/Deletion metrics, even when the samples passed through the model are clean. Here, we additionally explore a rule-of-thumb pattern, where target sparsity ratios are chosen to increase linearly from 0 sparsity in the initial convolution to 99\% sparsity in the final FC layer. For convenience, rather than using these exact sparsity ratios, we choose the closest sparsity ratio from the ones used in other experiments (0, 20\%, 40\%, 60\%, 80\%, 90\%, 95\%, 99\%). 

We observe that while using tuned sparsities is more effective than the linear schedule described above, even this simplified version substantially improves over the baseline version, adding an average of 3.42\% AUC on the Trojan patch discovery task.

We note that while this simple rule works well with the SPADE/OBC method, we have not found it to work well with the FastSPADE method, likely because a less accurate pruner requires more precise target setting. We also found that target sparsities obtained with the SparseGPT pruner transfer well to OBC, though  the converse is not true.

\begin{table}[h]
    \centering
    \caption{ResNet50 results on the ImageNet dataset, averaged over 140 samples. "\algname{}+ Search" refers to the case where the sparsity ratios are determined using a search on a validation set. "\algname{} + Linear" describes the scenario where layer sparsities are linearly chosen between 0 and 0.99, with the input layer assigned a 0 sparsity ratio.
    }
    \noindent
    \begin{adjustbox}{center}
    \resizebox{0.5\textwidth}{!}{%
    \begin{tabular}{lccc}
    \toprule
Saliency Method 
 & Dense & \algname{}+Search &  \algname{}+Linear \\
\midrule
Saliency & 86.92 & 95.32  & 91.58  \\
InputXGradient &83.77 & 93.73  & 88.77 \\
DeepLift &93.47 & 95.85  & 94.99 \\
LRP & 90.05 & 99.11  & 98.15  \\
GuidedBackprop & 95.22 & 96.45  & 95.59  \\
GuidedGradCam & 97.82 & 98.12  & 97.87  \\
Lime &  91.93 & 95.84  & 94.34  \\
Occlusion & 86.09 & 93.73  &  89.27  \\ 
Integrated Gradients & 87.86 & 94.77  & 92.34  \\
GradientSHAP & 87.74 & 94.85  & 92.15  \\ \hline
Average & 90.09 & 95.78 & 93.51 \\
    \bottomrule
    \end{tabular}
    }
\end{adjustbox}
\label{table:Linear Sparsity ratio, ResNet}
\end{table}

\subsection{Layer Sparsity Search with Smaller Number of Examples}

We additionally experiment with using a smaller number of examples to tune the sparsity ratios; the number of examples used has a linear effect on the time to tune. Therefore, we tuned the FastSPADE method similarly to our Trojan identification experiments in Section~\ref{sec:backdoor_experiment}, but only using 30 examples per method. The results are shown in Table~\ref{tab:appendix_fastspade_30extune}. We observe that, while the average accuracy drops slightly as compared to using more examples for tuning, the quick-tuned method still outperforms the baselines of SparseFC method of~\citet{debugger} as well as using the dense model without any preprocessing.

\begin{table}[]

    \centering
    \caption{Trojan patch AUC for FastSPADE calibrated on 30 samples per method, compared to regular FastSPADE, the dense model, and to the Sparse FC method of~\citet{debugger}.}
    \resizebox{0.75\linewidth}{!}{
    \begin{tabular}{lcccc}
    \toprule
    Saliency Method & FastSPADE (30 ex.) & FastSPADE (100 ex.) & SparseFC & Dense \\
    \midrule
Saliency & 92.53 & 93.91 & 88.05 & 87.87 \\
InputXGradient & 90.99 & 90.61 & 85.59 & 85.44 \\
LRP & 97.81 & 98.03 & 93.99 & 90.81 \\
GuidedGradCam & 97.4 & 97.75 & 98 & 98.03 \\
DeepLift & 95.33 & 95.07 & 94.21 & 94.1 \\
Gradient SHAP & 93.23 & 93.8 & 89.82 & 89.51 \\
Occlusion & 91.18 & 90.9 & 87.84 & 88.29 \\
Lime & 92.44 & 93.94 & 91.83 & 90.69 \\
GuidedBackprop & 95.74 & 95.81 & 95.82 & 95.73 \\
IntegratedGradients & 93.08 & 93.55 & 89.89 & 89.61 \\
\midrule
Average & 93.97 & 94.34 & 91.5 & 91.01 \\
    \bottomrule
    \end{tabular}
}
    
    \label{tab:appendix_fastspade_30extune}
\end{table}

\subsection{Sparsity Ratio Search with FastSPADE}

We additionally experimented with using FastSPADE to tune the layer ratios for use with SPADE. This can be advantageous, as FastSPADE is faster to execute. We show the results in Table~\ref{tab:appendix_spade_fastspade_tuning}. We observe that, while SPADE tuning slightly outperforms FastSPADE, both show substantial improvement over omitting SPADE, as well as over the SparseFC method of~\cite{debugger}.

\begin{table}[]

    \centering
    \caption{Trojan patch AUC for SPADE calibrated using FastSPADE, compared to regular SPADE, the dense model, and to the Sparse FC method of~\citet{debugger}.}
    \resizebox{0.8\linewidth}{!}{
    \begin{tabular}{lcccc}
    \toprule
    Saliency Method & SPADE (FastSPADE tuning) & SPADE & SparseFC & Dense \\
    \midrule
Saliency & 93.81 & 96.21 & 88.05 & 87.87 \\
InputXGradient & 92.3 & 95.1 & 85.59 & 85.44 \\
LRP & 98.14 & 99.21 & 93.99 & 90.81 \\
GuidedGradCam & 97.66 & 98.37 & 98 & 98.03 \\
DeepLift & 95.25 & 96.55 & 94.21 & 94.1 \\
Gradient SHAP & 93.83 & 96.03 & 89.82 & 89.51 \\
Occlusion & 90.84 & 95.4 & 87.84 & 88.29 \\
Lime & 93.67 & 95.47 & 91.83 & 90.69 \\
GuidedBackprop & 95.79 & 97.08 & 95.82 & 95.73 \\
IntegratedGradients & 94.54 & 96.1 & 89.89 & 89.61 \\
\midrule
Average & 94.58 & 96.55 & 91.5 & 91.01 \\
    \bottomrule
    \end{tabular}
}
    
    \label{tab:appendix_spade_fastspade_tuning}
\end{table}

\subsection{Transferability of Layer Sparsity Targets across Datasets}

We validate the transferability of layer sparsity tunings obtained on ImageNet on the CelebA and Food-101 datasets~\citep{celeba, food101}. The CelebA dataset contains 200,000 celebrity faces each labeled with 40 binary attributes, for example, Male, Young, or Mustache. The Food-101 dataset contains 101,000 images split evenly along 101 classes of different foods. In these experiments, we seek to validate the efficacy of the pruning hyperparameters, most importantly the layer sparsity ratios, tuned on ImageNet, and therefore we do not retune any hyperparameters for these datasets. Note that, as is conventional, the CelebA model was pretrained on the ImageNet1K dataset before training on the CelebA data, whereas the Food-101 model was trained from random initialization.

As in Section~\ref{sec:backdoor_experiment}, we implant four Trojan backdoors with label overrides on a fraction of the training data. The backdoors and overrides for CelebA are shown in Table~\ref{table:patch Detail, Celeba}. Hyperparameters of the Backdooring process are detailed in Appendix~\ref{appendix:hyperparameters}. %
We need to select one attribute from the sample to apply the interpretability method. Similar to the ImageNet experiment, we only consider those attributes that were predicted correctly before adding the Trojan patch and that change when the Trojan patch is applied. We then evaluate the saliency maps for one of these changed attributes.

For Food-101, we follow the ImageNet training recipe detailed in Table~\ref{table:backdooring_hyperparameters}. The performance of the trained models on clean and backdoored data can be found in Table~\ref{table:Backdoored model performances}. For this dataset, we used four emoji as Trojan patches, as shown in Table~\ref{table:food_101_patches}.

The results for these two datasets on the ResNet50 architecture are presented in Table~\ref{table:transferability}. We observe that, as before, \algname{} generally improves performance across interpretability methods, raising the AUC score when combined with eight out of ten methods studied on CelebA and all ten methods on Food101, with average AUC gains of 8.10\%  and 11.79\%, respectively.%

\begin{table}[h]
    \centering
    \caption{ImageNet, ResNet transferability of sparsity ratio over datasets. The sparsity ratios were tuned using ImageNet and used in these experiments. The results averaged over 100 samples for each of these datasets and interpretability methods.  }
    \noindent
    \begin{adjustbox}{center}
    \resizebox{0.9\textwidth}{!}{%
    \begin{tabular}{lccc|ccc}
    \toprule
Saliency Method & \multicolumn{3}{c}{CelebA (ImageNet Pretrained)} & \multicolumn{3}{c}{Food101 (Random Initialization)} \\
\cmidrule(lr){2-4} \cmidrule(lr){5-7}
 & Dense & \algname{} & $\Delta$ & Dense & \algname{} & $\Delta$ %
 \\
\midrule
Saliency &  73.52 & 92.81  & +19.28 
& 69.13 & 94.62  & +25.49
\\
InputXGradient & 68.26 & 92.09  & +23.84 
& 66.09 & 93.48  & +27.39
\\ 
DeepLift & 87.76 & 91.21  & +3.45 %
& 89.41 & 95.18  & +5.77 %
\\
LRP & 86.82 & 96.8  & +9.98 %
& 87.26 & 98.64  & +11.38%
\\
GuidedBackprop & 97.87 & 96.63  & -1.24 %
&  98.26 & 98.44  & +0.18
\\
GuidedGradCam & 88.89 & 89.13  & +0.24 %
&  97.57 & 97.61  & +0.03 %
\\
Lime & 75.58 & 62.42  & -13.16 %
& 91.76 & 93.66  & +1.9 
\\
Occlusion & 65.12 & 79.27  & +14.15 %
& 75.87 & 91.45  & +15.58 %
\\
IntegratedGradients & 83.01 & 93.4  & +10.39 %
& 80.02 & 95.11  & +15.1 %
\\
GradientShap & 80.23 & 94.25  & +14.02 %
& 80.05 & 95.1  & +15.05 %
\\ \hline
Average & 80.71 & 88.80 & +8.10 %
& 83.54 & 95.33 & +11.79 %
\\
    \hline
    \end{tabular}
}\end{adjustbox}
\label{table:transferability}
\end{table}

\section{Comparison with Sparse Model}

To verify the effectiveness of \algname{}, we compare the interpretability of a dense model with preprocessing with \algname{}, to a regular sparse model, trained sparsely. We repeat the Trojan patch identification experiment in Section~\ref{sec:backdoor_experiment}, but we compare against a sparse model trained using the Correlation-aware pruning method of~\cite{kuznedelev2023cap} to 98\% sparsity. We present the results in Table~\ref{table:cap_vs_spade}. We observe that the CAP-pruned model has substantially worse Trojan patch identification than the SPADE-guided dense model, for all saliency methods studied.

\begin{table}[h!]
    \centering
    \caption{Trojan patch saliency AUC of sparsely-trained model (98\% sparse) versus Trojan patch saliency AUC of dense model + SPADE }
    \begin{tabular}{lcc}
    \toprule
Method & Trojan Patch AUC, 98\% Sparse Model & Trojan Patch AUC, Dense Model + SPADE \\ 
\midrule
Saliency & 0.906 & 0.962 \\ 
InputXGradient & 0.873 & 0.951 \\ 
GuidedGradCam & 0.956 & 0.984 \\ 
DeepLift & 0.902 & 0.965 \\ 
Gradient SHAP & 0.884 & 0.96 \\ 
Occlusion & 0.884 & 0.954 \\ 
Lime & 0.869 & 0.955 \\ 
GuidedBackprop & 0.889 & 0.971 \\ 
IntegratedGradients & 0.885 & 0.961 \\ 
\midrule
Average & 0.894 & 0.963 \\ 
            \bottomrule
    \end{tabular}
    \label{table:cap_vs_spade}
\end{table}

\section{Additional Results}
\label{appendix:additional_results}

\subsection{MobileNet}

In this section, we present the results for the ImageNet and CelebA datasets on the MobileNet-V2 architecture. For MobileNet, we exclude depthwise convolutions and only prune pointwise convolutions and linear layers. Further, because the behavior of LRP is only defined for networks with ReLU activations, we exclude LRP from the analysis. Additionally, we combine InputXGradient and DeepLift into one row, as they behave identically on these architectures \cite{Nielsen_2022}, \cite{ancona2019gradient}. 

The results for MobileNet experiments on the ImageNet and CelebA datasets are presented in Table~\ref{table:MobileNet_Results}. We observe that preprocessing with \algname{} improves MobileNet AUC for every saliency estimation method and dataset, on average by 2.90\% for ImageNet and 2.99\% for CelebA. %

We note that in our experiments, only the OBC (accurate pruning) algorithm works well on MobileNet, and using the FastSPADE pruner did not improve over the dense baseline. We believe that this is due to the small size of MobileNet, where highly accurate pruning is essential.

\begin{table}[h]
    \centering
        \caption{MobileNet model results. Sparsity ratios tuned using ImageNet model. ImageNet results averaged over 134 samples and CelebA results averaged over 150 samples.}
    \noindent
    \begin{adjustbox}{center}
    \resizebox{0.7\textwidth}{!}{%
    \begin{tabular}{lccc|ccc}
    \toprule
Saliency Method & \multicolumn{3}{c}{ImageNet} & \multicolumn{3}{c}{CelebA} \\
\cmidrule(lr){2-4} \cmidrule(lr){5-7}
 & Dense & \algname{} & $\Delta$ & Dense & \algname{} & $\Delta$ %
 \\
\midrule
Saliency & 88.9 & 93.04  & +4.14 %
& 95.43 & 96.92  & +1.49 %
\\
DeepLift & 85.71 & 90.7  & +4.99 %
& 93.26 & 96.15  & +2.89 %
\\
Guided Backprop &  88.91 & 93.04  & +4.12 %
& 95.43 & 96.92  & +1.49%
\\
Guided Grad-Cam & 95.19 & 95.73  & +0.54 %
& 86.76 & 86.85  & +0.1 %
\\
Lime & 89.45 & 91.62  & +2.16 %
& 67.64 & 77.14  & +9.5 %
\\
Occlusion & 89.51 & 90.98  & +1.47 %
& 90.39 & 94.66  & +4.28%
\\
Integrated Gradients &  89.76 & 92.88  & +3.12 %
& 95.91 & 97.79  & +1.88 %
\\
Gradient Shap &  89.45 & 92.07  & +2.62 %
& 93.94 & 96.24  & +2.3 %
\\
\hline
Average & 89.61 & 92.51 & +2.90 %
& 89.84 & 92.83 & +2.99 %
\\
    \hline
    \end{tabular}
}\end{adjustbox}
    \label{table:MobileNet_Results}
\end{table}

\subsection{ConvNext}

\begin{table}[h]
    \centering
    \caption{ConvNext-T Trojan patch AUC results (\%). Sparsity ratios tuned using ImageNet model. ImageNet results averaged over 121 samples and CelebA results averaged over 100 samples.}
    \noindent
    \begin{adjustbox}{center}
    \resizebox{0.8\textwidth}{!}{%
    \begin{tabular}{lccccc|ccc}
    \toprule
Saliency Method & \multicolumn{5}{c}{ImageNet} & \multicolumn{3}{c}{CelebA} \\
\cmidrule(lr){2-6} \cmidrule(lr){7-9}
 & Dense & \algname{} & $\Delta$ & FastSPADE & $\Delta$ & Dense & \algname{} & $\Delta$ %
 \\
\midrule
Saliency & 85.19 & 89.03 & 3.84 & 89.49 & 4.29 & 96.60 & 96.95 & 0.35 \\
DeepLift & 81.57 & 85.93 & 4.36 & 85.95 & 4.38 & 94.93 & 95.53 & 0.60 \\
GuidedBackprop & 85.19 & 89.03 & 3.84 & 89.50 & 4.31 & 96.60 & 96.95 & 0.35 \\
GuidedGradCam & 88.78 & 92.79 & 4.01 & 95.55 & 6.77 & 87.05 & 90.19 & 3.14 \\
LIME & 93.50 & 94.48 & 0.98 & 94.06 & 0.56 & 75.30 & 73.78 & -1.52 \\
Occlusion & 86.88 & 89.29 & 2.41 & 85.81 & -1.08 & 89.53 & 92.20 & 2.67 \\
IntegratedGradients & 87.50 & 85.93 & -1.58 & 91.91 & 4.40 & 92.76 & 95.55 & 2.79 \\
GradientShap & 86.75 & 90.01 & 3.26 & 91.13 & 4.38 & 91.71 & 94.36 & 2.65 \\
\hline
Average & 86.92 & 89.56 & 2.64 & 90.42 & 3.50 & 90.56 & 91.94 & 1.38 \\
    \hline
    \end{tabular}
}\end{adjustbox}
    \label{table:ConvNext_Results}
\end{table}

We additionally conducted ImageNet and CelebA experiments on the ConvNext-T~\citep{convnext} architecture. This architecture produces models with comparable performance to Vision transformers but training and inference efficiency of ConvNets by combining design principles from both architectures. Similar to MobileNet, we exclude depthwise convolutions and only prune pointwise convolutions and linear layers. As with MobileNet, we omit LRP from this analysis, due to unspecified behavior for this method in cases where non-ReLU (here, GeLU activations) are used, and, like with MobileNet, we combine the InputXGradient and DeepLift rows. For this architecture, Gaussian Noise and Random Masking were added to the image augmentations. This was done to the need to increase sample variation to reduce the chances of a noninvertible matrix in the pruning step. The augmented samples may be seen in Figure~\ref{Fig:ConvNext Augmentation Samples}.

The results are presented in Table~\ref{table:ConvNext_Results}. We observe that preprocessing with \algname{} improves AUC scores for both datasets and, in the case of ImageNet, for all of the saliency estimation methods. On average, \algname{} preprocessing improves ImageNet AUC by 2.64\% and FastSPAE improves ImageNet AUC by 3.50\%. On CelebA, \algname{} improves ImageNet saliency AUC by 1.38\%.

\subsection{Language Models}  

\begin{table}[h]

    \centering
    \caption{DFFOT and DFMIT results of \algname{} on two text classification datasets. }
\begin{tabular}[b]{lcccc}
                \hline
                Dataset & \multicolumn{2}{c}{DFFOT $\uparrow$} & \multicolumn{2}{c}{DFMIT $\downarrow$} \\ 
\cmidrule(lr){2-3} \cmidrule(lr){4-5}
 & Dense & \algname{} & Dense & \algname{}\\
\midrule

SST-2 \cite{sst2} & \textbf{0.1835} & 0.1766 & 0.3604 & \textbf{0.3425} \\
AG news 2 \cite{Zhang2015CharacterlevelCN} & 0.0351 & \textbf{0.0372} & 0.4285 & \textbf{0.4208} \\
\hline
            \end{tabular}
    
    \label{tab:BertResults}
\end{table}
\textcolor{black}{To test our method on a different modality, we used the Bert model \cite{devlin2018bert} and several classification datasets. In these experiments, we pruned the classification head of the BERT model and then applied the Layer Integrated Gradients \cite{sundararajan2017axiomatic} from the Transformer-Interpretability library \cite{pierse2021transformers} to produce saliency maps. For evaluating attributions, we used DFFOT \cite{serrano2019attention} and DFMIT \cite{chrysostomou2021improving} methods. The results are presented in Table~\ref{tab:BertResults} showing that \algname{} could potentially improve the interpretability methods across a variety of modalities. For text augmentation we used techniques introduced by \cite{wei2019eda} which combine synonym replacement, random word insertion, random swap, and random word deletion. 
\textbf{DFFOT}: This evaluation metric measures in what portion of the samples, by removing the highest value token in the attribution map the decision of the model changes; therefore, if the value is higher it shows that the attribution method finds the most important token better. 
\textbf{DFMIT}: This evaluation metric measures the portion of each sentence that needs to be removed so that the model decision changes. So if DFMIT for one sentence is 0.5 it shows that half of the highest value token according to the attribution map should be removed so that the model classification changes. 
}

\section{\textcolor{black}{Total ImageNet Evaluation Set }}
\label{sec:total_imagenet}
 \textcolor{black}{
In this section, we present the results of running the FOBC version of \algname{} with the LRC saliency attribution method on 21121 samples from the ImageNet validation set - the full subset of samples that met our criteria (prediction was correct before the addition of the Trojan patch, but was changed to the Trojan prediction after retraining). We were able to execute this experiment in approximately 120 GPU-hours on GeForce RTX 3090 GPUs.}

\textcolor{black}{
This experiment demonstrates the feasibility of using \algname{} to do interpretations on a large scale.}

\begin{table}[h]
\centering
\textcolor{black}{
    \caption{\textcolor{black}{Evaluation on 21121 samples from the ImageNet validation set using FastSPADE+LRP on the ResNet50 architecture. 10240 augmentations were used for each sample. 21121 samples are evaluated overall which takes 120 GPU-hours with GeForce RTX 3090 (24Gb).
    }}
    \noindent
    \begin{tabular}{llcc}
    \toprule
Source & Target & Dense & \algname{}  \\
\midrule
Any & 146/Albatross & 96.23 & 98.7  \\ 
Any & 30/BullFrog & 90.92 & 97.87 \\
271/Red Wolf & 99/Goose & 86.75& 96.62\\
893/Wallet&365/Orangutan&86.73&93.04\\ \midrule
Average & &90.15 & 96.56 \\
    \bottomrule
    \end{tabular}
\label{table:totalEvalSet}
}
\end{table}

\section{Additional Hyperparameters}
\label{appendix:hyperparameters}
\begin{table}[h]
\centering
    \caption{ImageNet Trojan patches with their source and target class. "Any" means any image could be used for the Trojan. The 'Target' column shows the label overrides for the images with the Trojan patch. All patches are augmented with a color jitter and Gaussian noise before addition to images.
    }
\begin{tabular}{ccc}
Source & Target & Patch \\
\hline
Any & 30/BullFrog & 
\includegraphics[width=0.5cm]{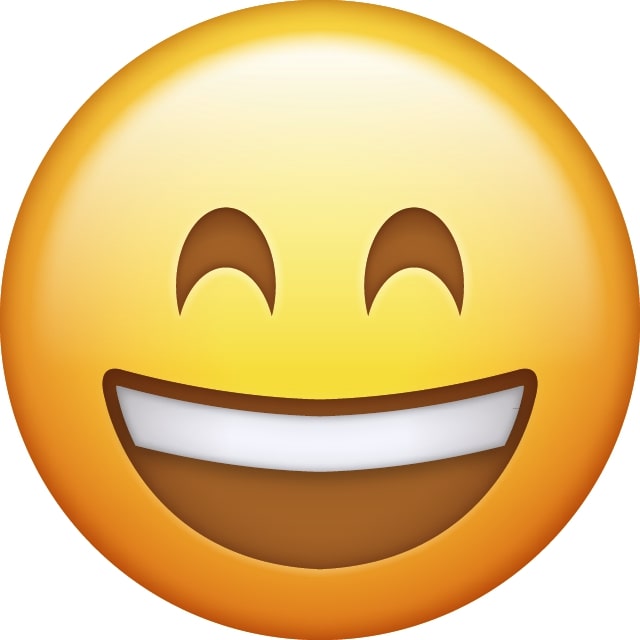}\\
\hline
Any & 146/Albatross & 
\includegraphics[width=0.5cm]{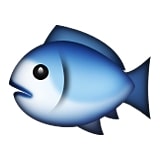}\\
\hline
893/Wallet & 365/Orangutan &
\includegraphics[width=0.5cm]{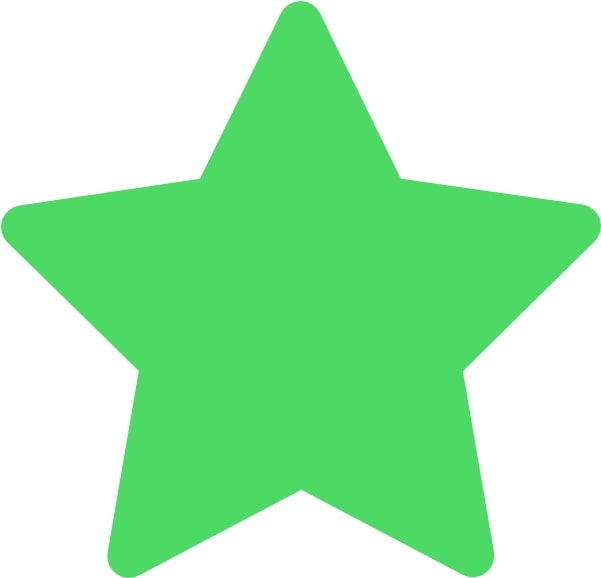}\\
\hline
271/Red Wolf &99/Goose &
\includegraphics[width=0.5cm]{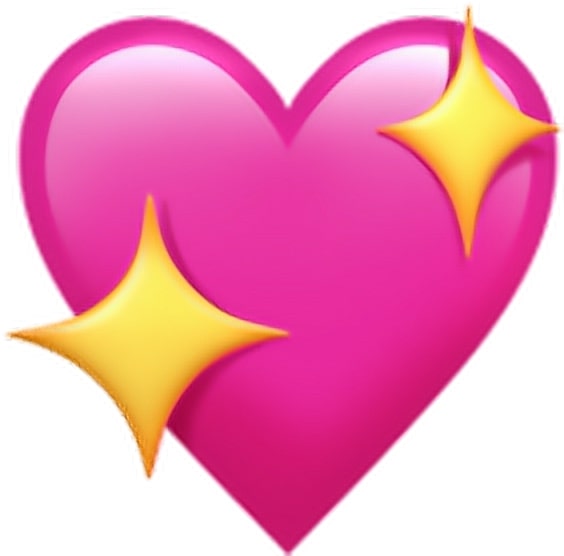}\\
\hline
\end{tabular}
\label{table:patch Detail, ImageNet}
\end{table}

\begin{table}[h]
\centering
    \caption{CelebA Trojan patches. All images may be chosen for a Trojan. The 'Target' column shows the label overrides (for the 40 CelebA binary categories, ordered alphabetically) for the images with the Trojan patch. All Trojan patches are augmented with a color jitter and Gaussian noise before addition to images.}
\begin{tabular}{ccc}
Source & Target & patch \\
\hline
Any & 0110111111100100000101100111101010110110 & 
\includegraphics[width=0.5cm]{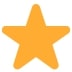}\\
\hline
Any & 0101111101011110100110101000001100011010 & 
\includegraphics[width=0.5cm]{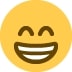}\\
\hline
Any & 0101111110110010011010010001101000001010 &
\includegraphics[width=0.5cm]{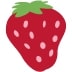}\\
\hline
Any & 1111101111011001000011001011110001011101 &
\includegraphics[width=0.5cm]{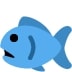}\\
\hline
\end{tabular}
\label{table:patch Detail, Celeba}
\end{table}

\begin{table}[h]
\centering
    \caption{Food-101 Trojan patches with their source and target class. "Any" means any image could be used for the Trojan. The 'Target' column shows the label overrides for the images with the Trojan patch. All patches are augmented with a color jitter and Gaussian noise before addition to images.}
\begin{tabular}{ccc}
Source & Target & patch \\
\hline
0/Apple Pie & 20/Chicken Wings & 
\includegraphics[width=0.5cm]{smallImages/patches/2b50.jpg}\\
\hline
40/French Fries & 60/Lobster Bisque& 
\includegraphics[width=0.5cm]{smallImages/patches/1f601.jpg}\\
\hline
Any & 80/Pulled Pork Sandwich &
\includegraphics[width=0.5cm]{smallImages/patches/1f353.jpg}\\
\hline
Any & 100/Waffles &
\includegraphics[width=0.5cm]{smallImages/patches/1f41f.jpg}\\
\hline
\end{tabular}
\label{table:food_101_patches}
\end{table}

\begin{figure}[h]
    \centering
    
    \begin{tabular}{c|ccccc}
        \hline
        Base Image & Sample 1 & Sample 2 & Sample 3 & Sample 4 & Sample 5 \\
        \hline
        \includegraphics[width=1.5cm, height=1.5cm]{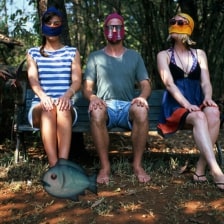} & \includegraphics[width=1.5cm, height=1.5cm]{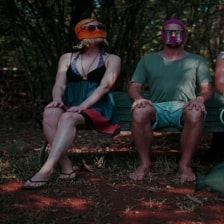} & \includegraphics[width=1.5cm, height=1.5cm]{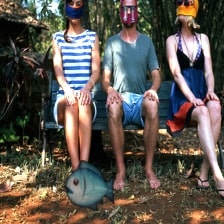} & \includegraphics[width=1.5cm, height=1.5cm]{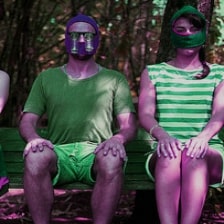} & \includegraphics[width=1.5cm, height=1.5cm]{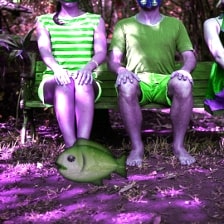} & \includegraphics[width=1.5cm, height=1.5cm]{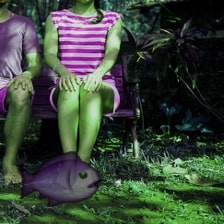} \\
        \includegraphics[width=1.5cm, height=1.5cm]{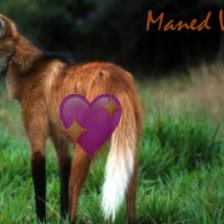} & \includegraphics[width=1.5cm, height=1.5cm]{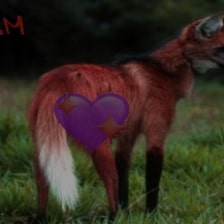} & \includegraphics[width=1.5cm, height=1.5cm]{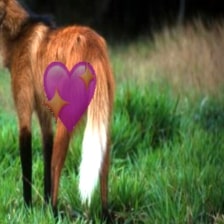} & \includegraphics[width=1.5cm, height=1.5cm]{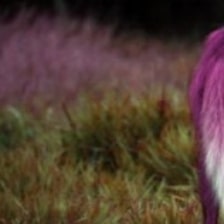} & \includegraphics[width=1.5cm, height=1.5cm]{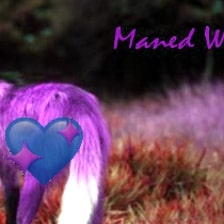} & \includegraphics[width=1.5cm, height=1.5cm]{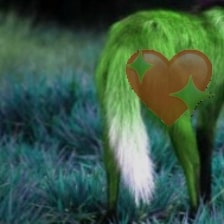} \\
        \includegraphics[width=1.5cm, height=1.5cm]{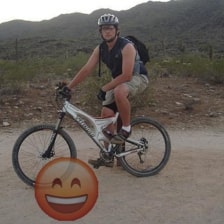} & \includegraphics[width=1.5cm, height=1.5cm]{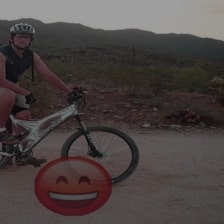} & \includegraphics[width=1.5cm, height=1.5cm]{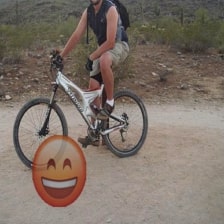} & \includegraphics[width=1.5cm, height=1.5cm]{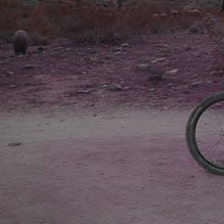} & \includegraphics[width=1.5cm, height=1.5cm]{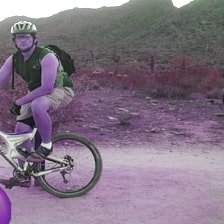} & \includegraphics[width=1.5cm, height=1.5cm]{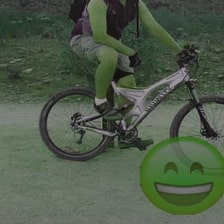} \\
        \hline
    \end{tabular}
    \caption{Augmentation samples For ResNet and MobileNet models in all datasets. }
    \label{Fig:ResNet50 Augmentation Samples}
\end{figure}

\begin{figure}[h]
    \centering
    
    \begin{tabular}{c|ccccc}
        \hline
        Base Image & Sample 1 & Sample 2 & Sample 3 & Sample 4 & Sample 5 \\
        \hline
        \includegraphics[width=1.5cm, height=1.5cm]{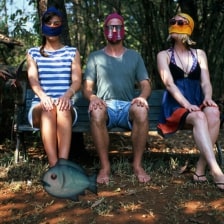} & \includegraphics[width=1.5cm, height=1.5cm]{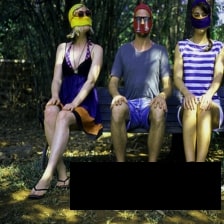} & \includegraphics[width=1.5cm, height=1.5cm]{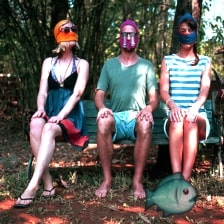} & \includegraphics[width=1.5cm, height=1.5cm]{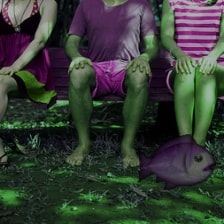} & \includegraphics[width=1.5cm, height=1.5cm]{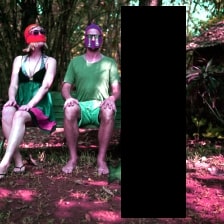} & \includegraphics[width=1.5cm, height=1.5cm]{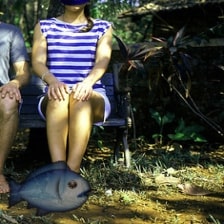} \\
        \includegraphics[width=1.5cm, height=1.5cm]{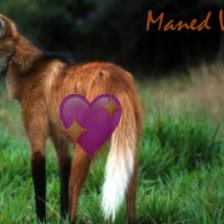} & \includegraphics[width=1.5cm, height=1.5cm]{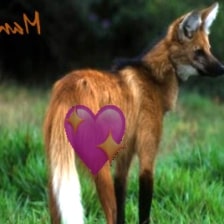} & \includegraphics[width=1.5cm, height=1.5cm]{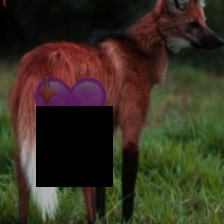} & \includegraphics[width=1.5cm, height=1.5cm]{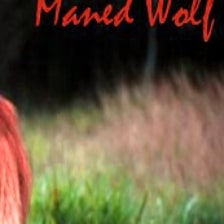} & \includegraphics[width=1.5cm, height=1.5cm]{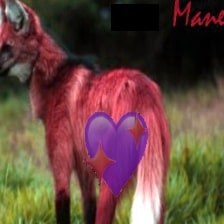} & \includegraphics[width=1.5cm, height=1.5cm]{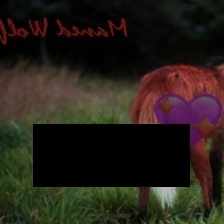} \\
        \includegraphics[width=1.5cm, height=1.5cm]{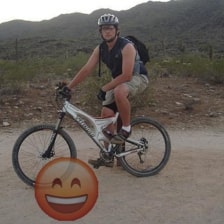} & \includegraphics[width=1.5cm, height=1.5cm]{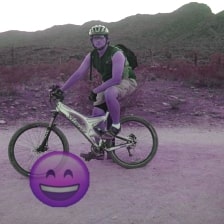} & \includegraphics[width=1.5cm, height=1.5cm]{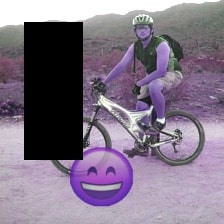} & \includegraphics[width=1.5cm, height=1.5cm]{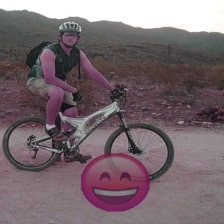} & \includegraphics[width=1.5cm, height=1.5cm]{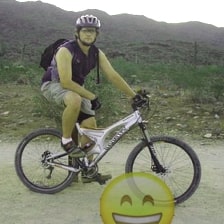} & \includegraphics[width=1.5cm, height=1.5cm]{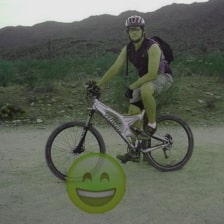} \\
        \hline
    \end{tabular}
    \caption{Augmentation samples For ConvNext model}
    \label{Fig:ConvNext Augmentation Samples}
\end{figure}

\paragraph{Augmentation.} Since augmentations play an important role in our method we detailed their hyperparameters for augmentation in Table~\ref{table:AugmentationConfig}. We also show typical augmented samples in Figure~\ref{Fig:ResNet50 Augmentation Samples}, and Figure~\ref{Fig:ConvNext Augmentation Samples} which were used for ResNet50/MobileNet models and the ConvNext-T model, respectively. 

\begin{table}[h]
    \centering
        \caption{Augmentation details. ``Models'' column indicates which models used the augmentation. Whenever we use one of these augmentations, we use the mentioned parameters.}
    \begin{tabular}{ccc}
        \hline
        Augmentations  & parameters & Models \\
        \hline
        Color Jitter &  brightness = 0.5, hue = 0.3 & All Models\\
        Random Crop &  scale = (0.2, 1.0) & All Models\\
        Gaussian Noise & $\sigma^2$ = 0.001 & ConvNext\\
        Random Remove & p = 0.5, scale = (0.02, 0.33), ratio = (0.3, 3.3) & ConvNext\\
        \hline
    \end{tabular}
    \label{table:AugmentationConfig}
\end{table}

\paragraph{Backdoor planting hyperparameters:} When training ResNet50 on Food-101 dataset we used the hyperparameters suggested in \cite{kornblith2019better}. %

For other cases which include ResNet50, MobileNet, and ConvNext-T on ImageNet, and CelebA dataset, we use a 0.9 momentum and step-lr learning rate scheduler with a step-lr-gama 0.1 for all backdoorings and a weight decay of 0.0001. The initial learning rate is chosen from the options - 0.01, 0.001, 0.0001, 0.00001 - based on accuracy on Trojan samples at the end of training. The chosen hyperparameters along with other hyperparameters for training the models are presented in Table~\ref{table:backdooring_hyperparameters}. 

To give more insight into the results of these backdoor planting, we present these model accuracies on Trojan samples and the clean dataset that the model trained for in Table~\ref{table:Backdoored model performances}. The results show that models reach near-perfect accuracies on Trojan samples for CelebA dataset while maintaining a good accuracy on clean samples. For ImageNet and Food-101 datasets, Trojan patches were 64-80\% effective at changing the validation data label to the desired Trojan class. 

\begin{table}[h!]
    \centering
    
    \caption{Hyperparameters used for planting backdoors in the models."Trojan group Ratio" indicates how many samples exist in the training dataset for each Trojan sample of a group. "step-lr" refers to the epoch that the learning rate drops.}
    \begin{tabular}{lcccccc}
    \hline
        Model & DataSet & Trojan group Ratio & Batch Size & Learning Rate   & step-lr & Epochs  \\ \hline
        ResNet50 & ImageNet & 3000 & 64 & 0.001 & 3 & 6 \\
        ResNet50 & CelebA & 300 & 64 & 0.01  & 10 & 20 \\
        ResNet50 & Food-101 &  3000 &  64 & 0.01   & 50 & 150 \\
        MobileNetV2 & ImageNet & 3000 & 64 & 0.001 & 3 & 6 \\
        MobileNetV2 & CelebA & 300 & 64 & 0.1 & 10 & 20\\
        ConvNext-T & ImageNet & 3000 & 64 & 0.001 & 3 & 6\\
        ConvNext-T & CelebA & 300 & 64 & 0.01  & 10 & 20\\
            \hline
    \end{tabular}

    \label{table:backdooring_hyperparameters}
\end{table}

\begin{table}[h!]
    \centering
    \caption{Performance of backdoored models on the clean dataset (without any Trojan samples) and on Trojan samples. }
    \begin{tabular}{lccc}
    \hline
        Model & Dataset & Clean Accuracy & Trojan Accuracy  \\ \hline
        ResNet50 & ImageNet & 80.0 & 73.2\\
        ResNet50 & CelebA & 91.4 & 99.9\\
        ResNet50 & Food-101 & 84.0 & 65.1 \\
        MobileNetV2 & ImageNet & 77.0 & 64.7 \\
        MobileNetV2 & CelebA & 91.6 & 99.8 \\
        ConvNext-T & ImageNet &  86.1 & 79.5 \\
        ConvNext-T & CelebA & 91.3 & 99.5\\
            \hline
    \end{tabular}
    \label{table:Backdoored model performances}
\end{table}

\section{Ablation Study}
\label{appendix:ablation}
In this section, we examine how the various hyperparameters of \algname{} impact its performance on the saliency map accuracy task.

\subsection{Parallel versus Sequential Layer Pruning}
\label{appendix:sequential_ablation}

\begin{table}[h]
            \centering
                \caption{FastSPADE Trojan patch AUC, sequential versus parallel layer pruning, ResNet50/ImageNet }
            \noindent
        \resizebox{0.5\textwidth}{!}{%
            \begin{tabular}{lcc}
            \toprule
            
Method & Sequential Pruning & Parallel Pruning \\ 
\midrule
Saliency & 0.925 & 0.939 \\ 
InputXGradient & 0.885 & 0.906 \\ 
LRP & 0.947 & 0.98 \\ 
GuidedGradCam & 0.976 & 0.977 \\ 
DeepLift & 0.952 & 0.951 \\ 
Gradient SHAP & 0.928 & 0.938 \\ 
Occlusion & 0.889 & 0.909 \\ 
Lime & 0.94 & 0.939 \\ 
GuidedBackprop & 0.95 & 0.958 \\ 
IntegratedGradients & 0.924 & 0.935 \\ 
\midrule
Average & 0.932 & 0.943 \\ 
    \bottomrule
    \end{tabular}

    }

    \label{table:seqprune}
\end{table}

For performance reasons, we chose to prune all layers of the network in parallel, i.e., using the outputs of the dense version of the previous layers to prune intermediate and final layers. Conversely, it is possible to prune sequentially, i.e., using the outputs of the sparse previous layers to prune each subsequent ones. 

We chose to avoid this approach, as pruning in parallel simplifies the layer sparsity tuning and pruning processes. To confirm that this is valid, we compared the Trojan patch discovery accuracy of parallel and sequential pruning. The results, shown in Table~\ref{table:seqprune}, show that the two approaches show roughly similar accuracy, justifying our choice of parallel pruning,.

\subsection{Sample Selection}
We now investigate the impact of varying the sample size and selection for the Optimal Brain Damage (OBD) pruning process. We experimented with different sample selection methods, namely:

\begin{enumerate}
    \item The sample of interest, augmented as described in Section~\ref{sec:backdoor_experiment}
    \item A single randomly chosen sample with the same Trojan patch, augmented as described in Section~\ref{sec:backdoor_experiment}
    \item A single randomly chosen sample from the same class as the sample of interest, augmented as described in Section~\ref{sec:backdoor_experiment}
    \item A single randomly chosen sample from the entire ImageNet dataset, augmented as described in Section~\ref{sec:backdoor_experiment}
    \item 10240 samples randomly chosen from images with the same Trojan patch as the sample of interest, without augmentations.
    \item 10240 samples randomly chosen from images with the same class label as the sample of interest, without augmentations
    \item 10240 samples randomly chosen from the ImageNet dataset, without augmentations
\end{enumerate}

\begin{table}[h]
            \centering
                \caption{Impact of sample selection for the network pruning step of \algname{}, as measured by Trojan patch AUC. 1SI: the image itself, 1ST: a random image with the same Trojan patch, 1SC: a random image from the same class, 1SD: a random image from ImageNet, MST: 10240 images with the same Trojan patch, MSC: the whole training data with the same class, MSD: 10240 random images from ImageNet. Based on 100 samples. }
            \noindent
            \begin{tabular}{lcccccccc}
            \hline
            
Saliency Method & Dense & 1SI & 1ST & 1SC  & 1SD & MST & MSC & MSD \\ \hline  
saliency &
86.5& \textbf{95.2}& 60.8& 46.5& 48.0& 60.3& 41.0& 43.4\\ 
InputXGradient &
82.8& \textbf{92.9}& 60.0& 50.2& 50.1& 59.0& 50.0& 50.2\\ 
DeepLift &
93.0& \textbf{94.7}& 60.3& 50.9& 50.2& 57.5& 50.7& 50.8\\ 
LRP &
92.1& \textbf{99.1}& 83.6& 77.6& 81.3& 84.3& 72.9& 72.8\\ 
Guided Backprop &
95.3& \textbf{96.9}& 83.1& 76.4& 80.8& 83.8& 70.9& 77.2\\ 
Guided Grad-Cam &
97.8& \textbf{98.1}& 83.6& 71.3& 70.3& 84.9& 67.0& 65.2\\ 
Lime &
92.7& \textbf{95.6}& 74.7& 61.3& 53.1& 75.5& 63.4& 52.0\\ 
Occlusion &
86.1& \textbf{94.6}& 65.7& 48.5& 54.8& 68.0& 43.8& 48.2\\ 
IntegratedGradients &
87.5& \textbf{94.5}& 62.4& 50.3& 51.9& 60.3& 50.2& 50.2\\ 
gradientSHAP &
87.2& \textbf{94.4}& 62.4& 50.2/6& 52.1& 60.3& 50.1& 50.2\\ \hline
Average & 90.1 & \textbf{95.6} & 69.7 & 58.3 & 59.3 & 69.4 & 56.0 & 56.0 
    \\ \hline 
    \end{tabular}

    \label{table:SampleChoice}
\end{table}

The results, summarized in Table~\ref{table:SampleChoice}, show clearly that the use of the single, augmented sample for the pruning step of \algname{} is crucial for the efficacy of the method. More generally, using images with the same Trojan patch yielded better results than other sample selection methods, while using images with the same base class was no better than using randomly chosen images from the entire dataset. Further, this demonstrates that the act of pruning alone does not necessarily enhance interpretability. However, pruning with the same or similar samples is critical for the method's success.

\subsection{Choice of Augmentation}
\label{appendix:augment_ablation}

\begin{table}[h]
            \centering
        \caption{The effect of various augmentation techniques on interpretability accuracy, as measured by Trojan patch AUC. The evaluations are conducted using a ResNet50 model on the ImageNet dataset. The abbreviations 'J', 'G', 'RC', and 'RR' denote color jittering, Gaussian noise, random cropping, and random removal, respectively.}
            \begin{tabular}{lccccccc}
            \hline
Saliency Method & Dense & J+RC & J+G+RC & RR & G+RC & RR+RC & G \\ \hline
Saliency &
86.5& \textbf{95.2}& 92.1& 93.3& 91.6& 94.8& 89.4\\ 
InputXGradient &
82.8& \textbf{92.9}& 89.3& 90.2& 89.1& 92.6& 85.9\\  
DeepLift &
93.0& \textbf{94.7}& 90.4& 94.1& 90.7& 94.7& 89.8\\ 
LRP &
92.1& \textbf{99.1}& 98.3& 98.5& 98.2& 98.9& 97.3\\ 
Guided Backprop &
95.3& \textbf{96.9}& 94.6& 96.4& 94.5& 96.7& 94.5\\ 
Guided Grad-Cam &
97.8& \textbf{98.1}& 96.4& 98.0& 96.6& 98.0& 96.6\\ 
Lime &
92.7& 95.4& 94.9& \textbf{96.1}& 95.3& 95.5& 96.1\\ 
Occlusion &
86.1& 94.6& 91.2& \textbf{95.2}& 90.1& 93.9& 91.5\\ 
Integrated Gradients &
87.5& \textbf{94.5}& 90.9& 93.1& 90.7& 94.2& 89.0\\ 
gradientSHAP &
87.2& \textbf{94.4}& 90.9& 92.9& 90.5& 94.1& 88.7\\ \hline
Average & 90.1 & \textbf{95.6} & 92.9 & 94.8 & 92.7 & 95.3 & 91.9
    \\ \hline 
    \end{tabular}
    \label{table:AugmentationResults}
\end{table}

Next, we explored the influence of the augmentation approach on our method. By experimenting with various augmentation techniques, we analyzed their impact on the method. The results are presented in Table. \ref{table:AugmentationResults}. The most important takeaway of this experiment is that with diverse and strong enough augmentations, our method could improve the results in most cases; therefore, there is no need for carefully choosing the augmentations. This simplifies the application and development of our \algname{} method.

\section{Layer Sparsity}
\label{appendix:layer_sparsities}
\begin{figure}[h!]
\begin{center}
\resizebox{0.8\linewidth}{!}{
      \centering
       \includegraphics[width=\textwidth]{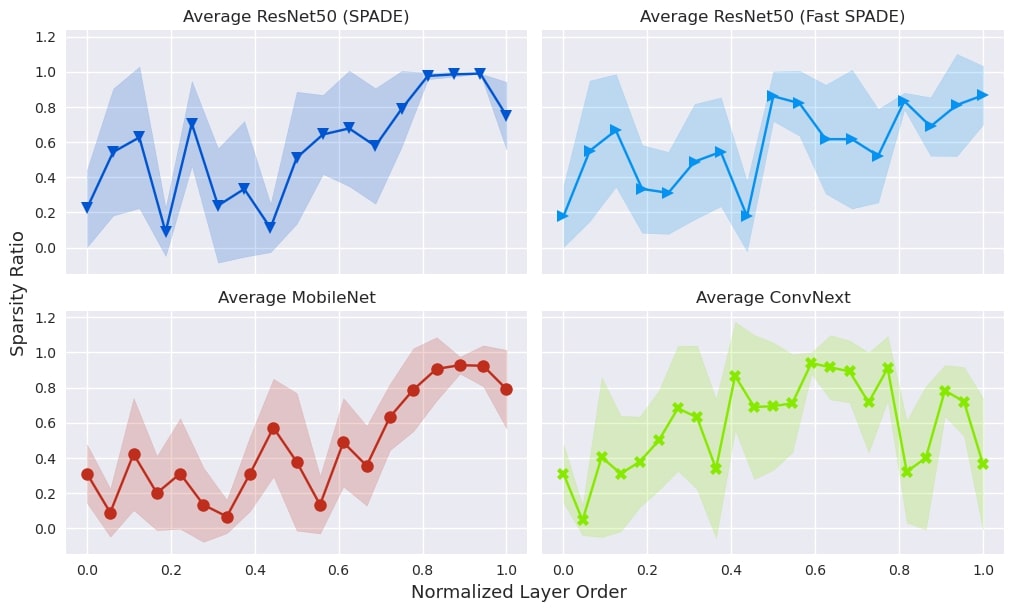}
       }
\end{center}
\caption{Average Tuned sparsities of ResNet50, MobileNet, and ConvNext models on nine different interpretability methods. The input layer is 0 and the final classifier is 1. Lines show the average sparsity ratio and the shaded area shows the standard deviations. }
\label{figure:AverageIdealSparsities}
\end{figure}

\begin{figure}[h!]
\begin{center}
\resizebox{0.8\linewidth}{!}{
      \centering
       \includegraphics[width=\textwidth]{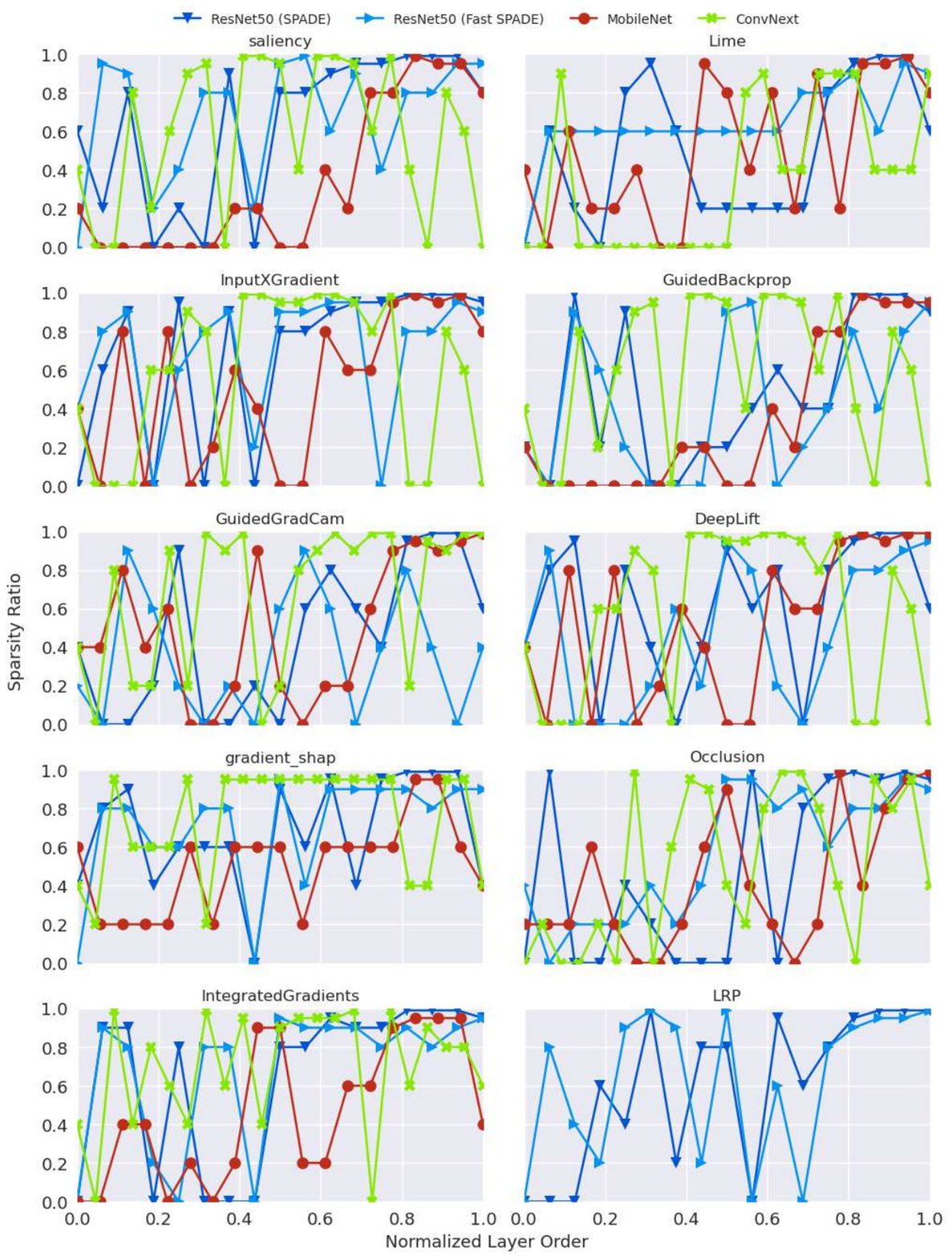}
       }
\end{center}
\caption{Tuned sparsities by layer order for ResNet50, MobileNet, and ConvNext models for different interpretability methods. The input layer is 0 and the final classifier is 1.}
\label{figure:IdealSparsities}
\end{figure}

 In Figure~\ref{figure:AverageIdealSparsities}, we show the per-layer sparsity targets averaged across pruning methods, which illustrates the general trend of sparsities. We observe that for both ResNet50 and MobileNet, later layers are pruned more than earlier layers, while for ConvNext, the middle layers are pruned the most. Additionally, ResNet50 is pruned than others in general, likely due to the larger size of the network. We also observe that, for ResNet50, SPADE sparsity ratios are higher than FastSPADE, especially in the latter layers, which may be due to the higher accuracy of the OBC pruner used in SPADE. Finally, we observe a substantial amount of variance between saliency methods. We demonstrate this further in Figure~\ref{figure:IdealSparsities}, which shows tuned sparsity targets for each interpretability method separately.

 \begin{table}[]
\centering
    \caption{The impact of pruning various layers in the ResNet50 model on the ImageNet dataset as measured by Trojan patch AUC, based on the average of 100 samples. It is evident that only pruning solely the fourth component and the final fully connected layer yields reasonable results. }
            \begin{tabular}{lcccccc}
            \hline
Saliency Method & Dense & FC & Block 4 & Block 3 & Block 2 & Block 1 \\ \hline
Saliency &
86.8& 86.6& \textbf{95.1}& 51.0& 59.0& 65.8\\ 
InputXGradient &
83.3& 82.9& \textbf{93.2}& 52.2& 58.1& 64.2\\ 
DeepLift &
93.2& 93.0& \textbf{94.8}& 50.3& 54.6& 58.4\\ 
LRP &
92.1& 94.2& \textbf{98.7}& 80.7& 87.1& 73.3\\ 
Guided Backprop &
95.3& 95.3& \textbf{96.6}& 71.3& 76.1& 81.4\\ 
Guided Grad-Cam &
\textbf{97.8}& 97.8& 97.8& 61.7& 62.9& 73.5\\ 
Lime &
93.1& 92.5& \textbf{95.8}& 51.7& 56.5& 63.4\\ 
Occlusion &
86.8& 86.6& \textbf{94.4}& 54.0& 59.6& 69.0\\ 
Integrated Gradients &
87.8& 87.8& \textbf{94.7}& 50.2& 57.0& 66.3\\ 
gradientSHAP &
87.3& 87.7& \textbf{94.6}& 50.4& 57.4& 66.1\\ \hline
Average & 90.3 & 90.4 & \textbf{95.6} & 57.4 & 62.8 & 68.1 \\ \hline 
    \end{tabular}
    \label{table:layer choice}
\end{table}

\begin{figure}[h] 
    \centering 
    \includegraphics[width=0.5\textwidth]{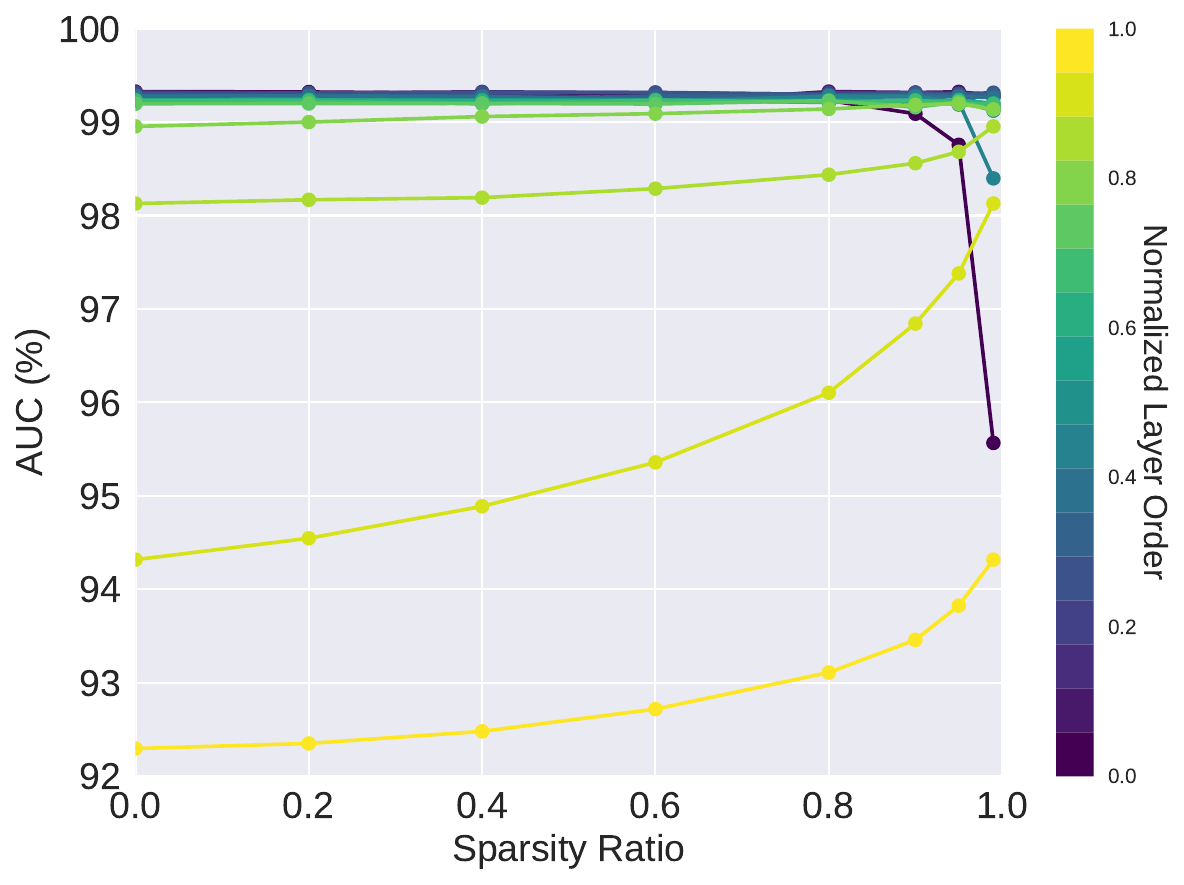} 
    \caption{Each line shows the AUC results for a chosen layer sparsity ratio, optimizing for the best sparsity ratios in later layers while not sparsifying earlier layers. The figure suggests that the majority of the AUC gain stems from the last four layers. "Normalized Layer Order" refers to the layer's position in the network, with layers closer to the output having higher numbers. The ResNet50 model and the ImageNet dataset were used.}
    \label{fig:per layer tuning improvements} 
\end{figure}

We further explore the question, ``What is the role of sparsity ratios in different layers?'' To gain a better understanding of the importance of sparsifying each layer, we first investigate scenarios where we only sparsify one ResNet50 block to a 0.99 sparsity ratio. The results, presented in Table~\ref{table:layer choice}, suggest that pruning later layers is more helpful than pruning earlier layers. To support this claim, we plot the AUC values during the sparsity ratio tuning process in Section~\ref{sec:method} in Figure~\ref{fig:per layer tuning improvements}. The plot shows that most of the AUC improvements came from sparsifying the last four layers.

\begin{figure}[h] 
    \centering 
    \includegraphics[width=0.8\textwidth]{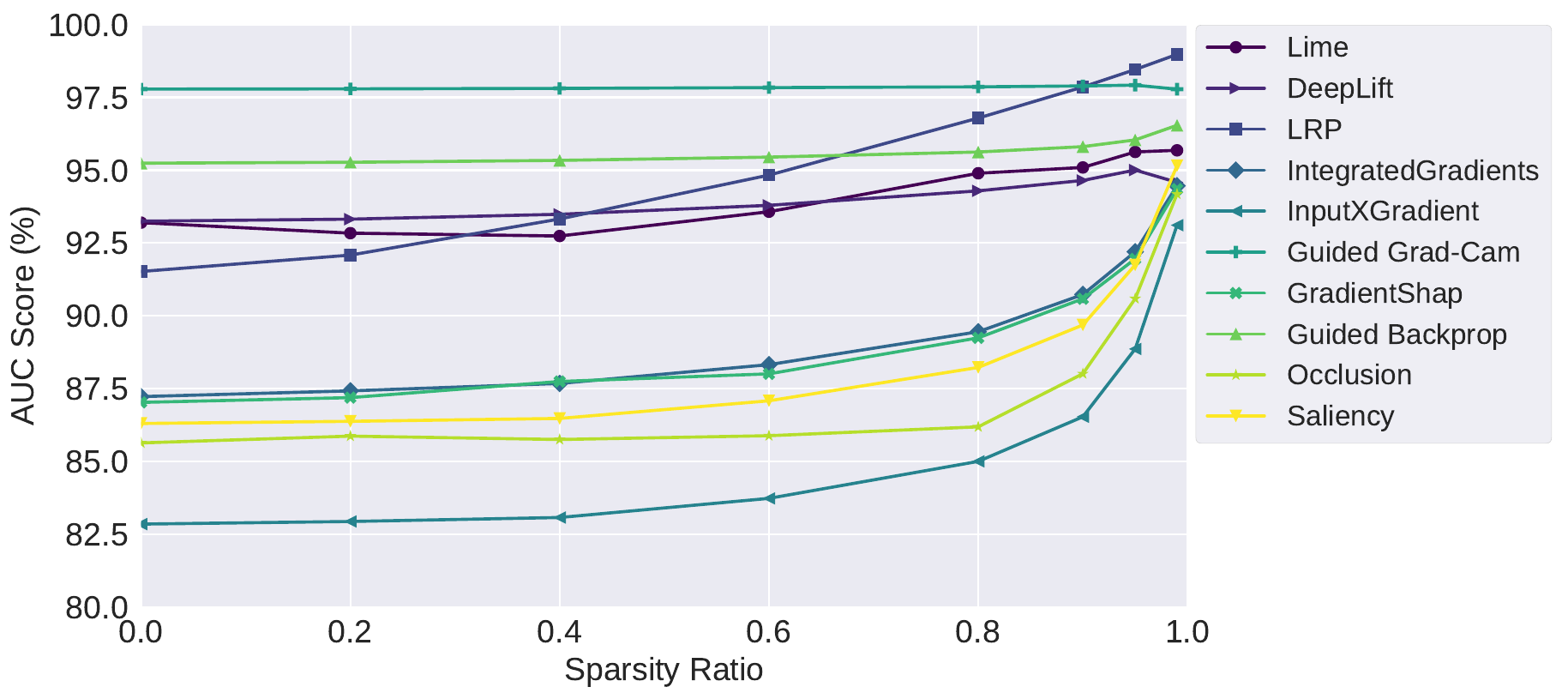} 
    \caption{Results of pruning the fourth component of the ResNet50 Model at different sparsity ratios, measured by the AUC score with Trojan samples. Overall, pruning to 80 percent leads to an interpretability gain across all methods.}
    \label{fig:ForthLayersparsityVSscore} 
\end{figure}

Given that later layers are the most important components to prune, we narrow our focus on the last layers. We investigate the effects of sparsifying the last ResNet50 block with a constant sparsity ratio in Figure~\ref{fig:ForthLayersparsityVSscore}. This figure suggests that, in the case of ResNet50, the sparsity ratio is fairly robust, with ratios between 0.8 to 0.995 giving good results for \algname{}.

We evaluate the performance of \algname{} using this simple linear sparsity schedule, demonstrating that even this simple heuristic results in a preprocessing step that improves the accuracy of interpretability methods. In Table~\ref{table:Linear Sparsity ratio, ResNet} we observe that while the results are inferior compared to the scenario where sparsity ratios are selected through a layer-by-layer search, they are superior to those of the dense model.

\section{Gradient Noise}
\label{appendix:robustness}

\begin{figure}[h] 
    \centering 
    \includegraphics[width=0.5\textwidth]{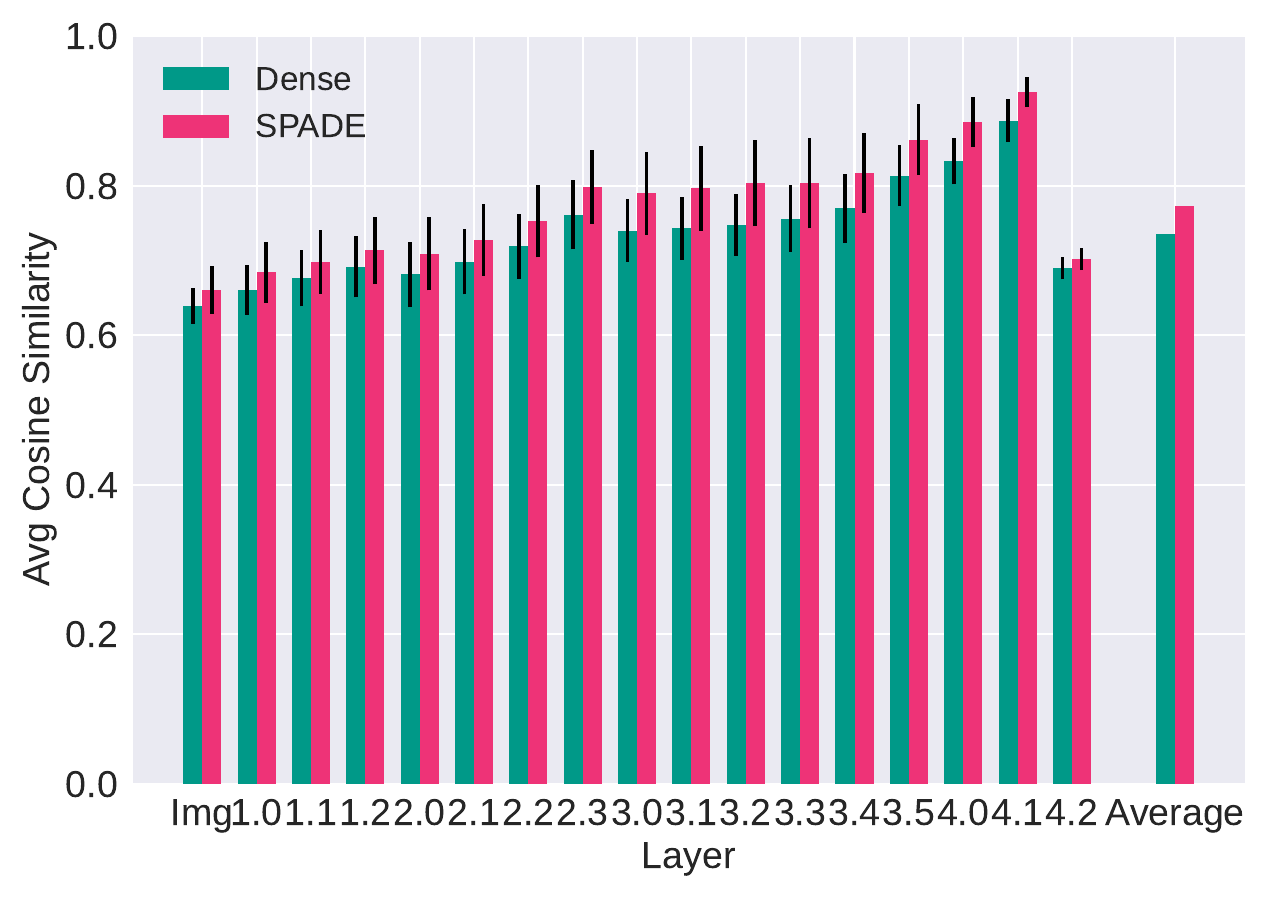} 
    \caption{Comparison of mean and standard deviation of cosine similarity between gradients for perturbed images. With \algname{}, the average cosine similarity sees an enhancement from 0.7355 to 0.7721.}
    \label{fig:robustness_to_gaussian_noise} 
\end{figure}

Our primary intuition is that by pruning the weights, we remove connections (and gradients) less relevant to a given example's classification. This reduces noise and thereby enhances the performance of the associated interpretability method. Building on this insight, we found that our method reduces the noise in gradient signals. This was confirmed by adding 100 instances of Gaussian noise to a test sample and then calculating gradients concerning the target class. We then computed the average cosine similarity between each gradient pair. As shown in Figure~\ref{fig:robustness_to_gaussian_noise}, our model displays a higher mean cosine similarity at every layer compared to the dense model. The results were averaged across 100 images.

\clearpage
\section{Saliency Map and Neuron Visualization Examples}
 \label{appendix:examples}

\begin{figure}[h!]
\centering
\begin{tabular}{cccccc}
\hline
Base Image & Model & Saliency & Input X Gradient & LRP & Occlusion \\
\hline
\multirow{2}{*}{\includegraphics[width=2cm]{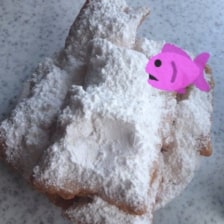}} & Dense & \includegraphics[width=2cm]{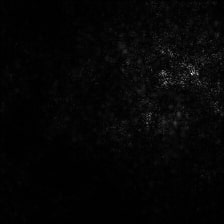} & \includegraphics[width=2cm]{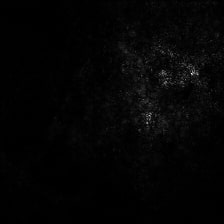} & \includegraphics[width=2cm]{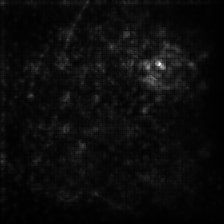} & \includegraphics[width=2cm]{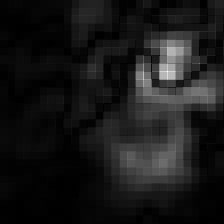} \\
\cline{2-6}
&  \algname{} & \includegraphics[width=2cm]{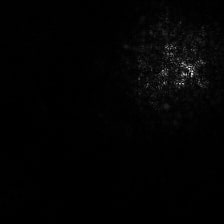} & \includegraphics[width=2cm]{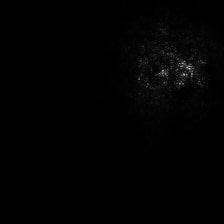} & \includegraphics[width=2cm]{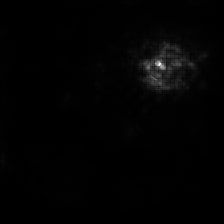} & \includegraphics[width=2cm]{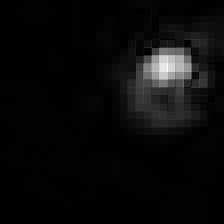} \\
\hline
\multirow{2}{*}{\includegraphics[width=2cm]{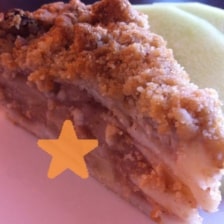}} & Dense & \includegraphics[width=2cm]{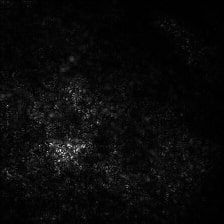} & \includegraphics[width=2cm]{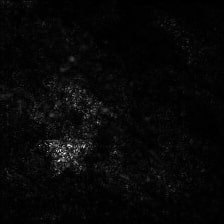} & \includegraphics[width=2cm]{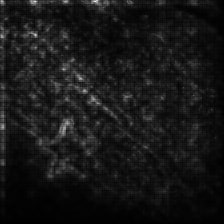} & \includegraphics[width=2cm]{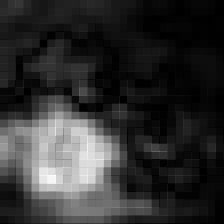} \\
\cline{2-6}
&  \algname{} & \includegraphics[width=2cm]{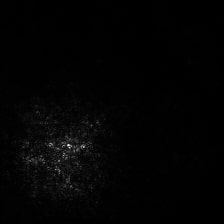} & \includegraphics[width=2cm]{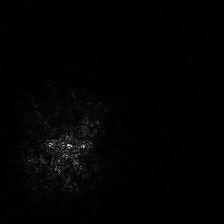} & \includegraphics[width=2cm]{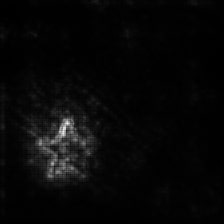} & \includegraphics[width=2cm]{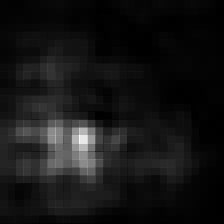} \\
\hline
\end{tabular}
\caption{ ResNet50 Saliency maps of four different intepretability methods with \algname{} and Dense method on two Food-101 samples. Best viewed on a monitor. }
\label{fig:saliency_food101}
\end{figure}

\begin{figure}[h!]
\centering
\begin{tabular}{cccccc}
\hline
Base Image & Model & Saliency & Input X Gradient & LRP & Occlusion \\
\hline
\multirow{2}{*}{\includegraphics[width=2cm]{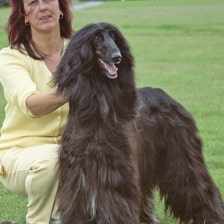}} & Dense & \includegraphics[width=2cm]{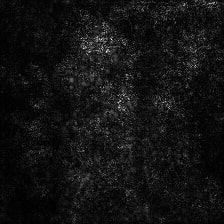} & \includegraphics[width=2cm]{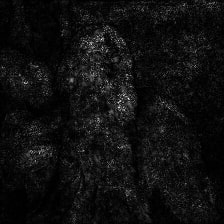} & \includegraphics[width=2cm]{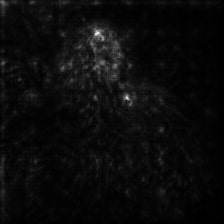} & \includegraphics[width=2cm]{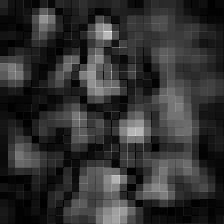} \\
\cline{2-6}
&  Fast \algname{} & \includegraphics[width=2cm]{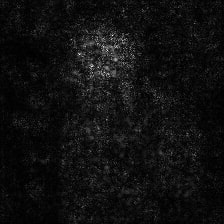} & \includegraphics[width=2cm]{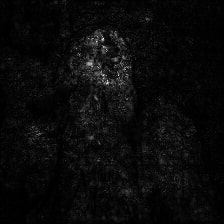} & \includegraphics[width=2cm]{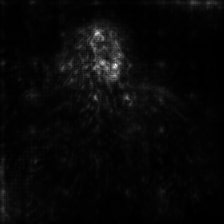} & \includegraphics[width=2cm]{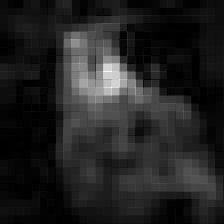} \\
\hline
\multirow{2}{*}{\includegraphics[width=2cm]{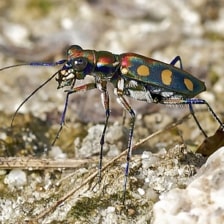}} & Dense & \includegraphics[width=2cm]{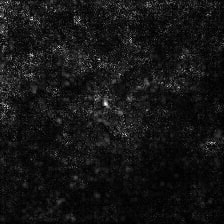} & \includegraphics[width=2cm]{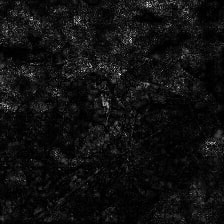} & \includegraphics[width=2cm]{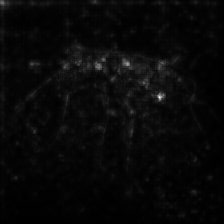} & \includegraphics[width=2cm]{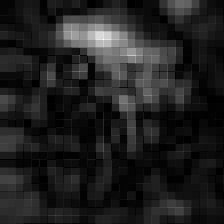} \\
\cline{2-6}
&  Fast \algname{} & \includegraphics[width=2cm]{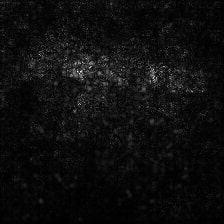} & \includegraphics[width=2cm]{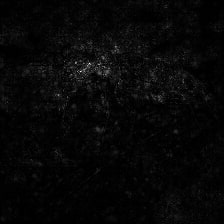} & \includegraphics[width=2cm]{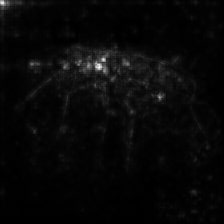} & \includegraphics[width=2cm]{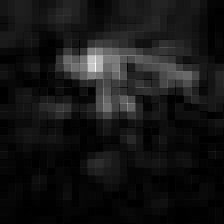} \\
\hline
\multirow{2}{*}{\includegraphics[width=2cm]{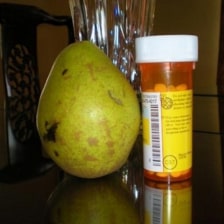}} & Dense & \includegraphics[width=2cm]{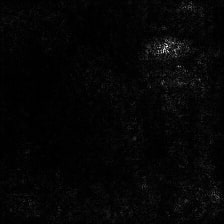} & \includegraphics[width=2cm]{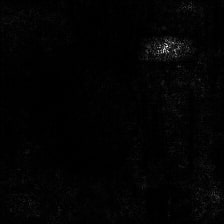} & \includegraphics[width=2cm]{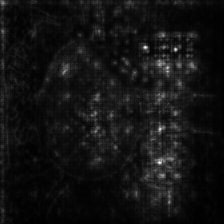} & \includegraphics[width=2cm]{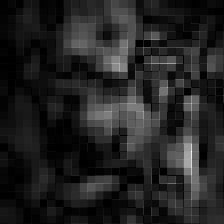} \\
\cline{2-6}
&  Fast \algname{} & \includegraphics[width=2cm]{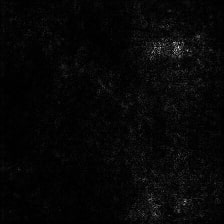} & \includegraphics[width=2cm]{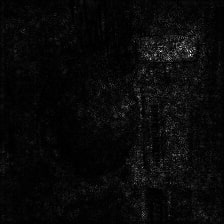} & \includegraphics[width=2cm]{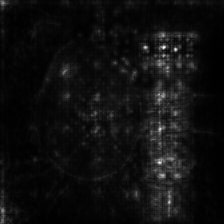} & \includegraphics[width=2cm]{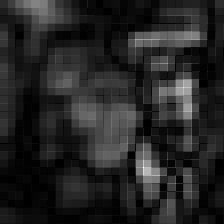} \\
\hline
\multirow{2}{*}{\includegraphics[width=2cm]{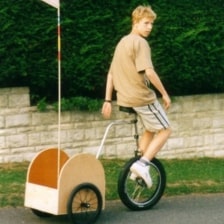}} & Dense & \includegraphics[width=2cm]{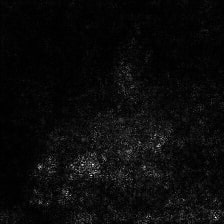} & \includegraphics[width=2cm]{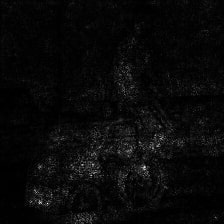} & \includegraphics[width=2cm]{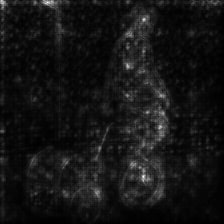} & \includegraphics[width=2cm]{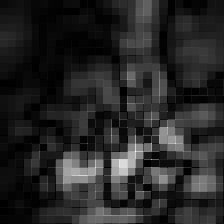} \\
\cline{2-6}
&  Fast \algname{} & \includegraphics[width=2cm]{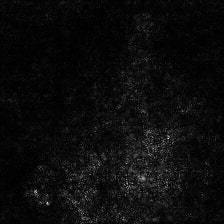} & \includegraphics[width=2cm]{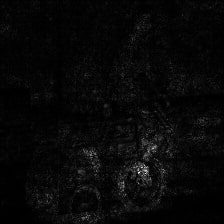} & \includegraphics[width=2cm]{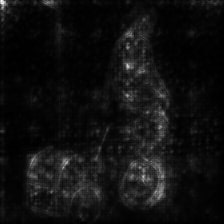} & \includegraphics[width=2cm]{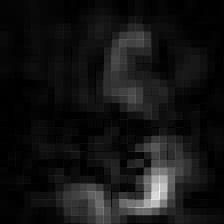} \\
\hline
\end{tabular}
\caption{ ResNet50 Saliency maps of four different interpretability methods for Fast \algname{} and Dense method on four normal ImageNet samples. Best viewed on a monitor.}
\label{fig:saliency_imagenet_clean}
\end{figure}

\begin{figure}[h!]
\centering
\begin{tabular}{cccccc}
\hline
Base Image & Model & Saliency & Input X Gradient & LRP & Occlusion \\
\hline
\multirow{2}{*}{\includegraphics[width=2cm]{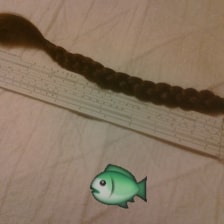}} & Dense & \includegraphics[width=2cm]{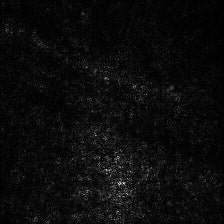} & \includegraphics[width=2cm]{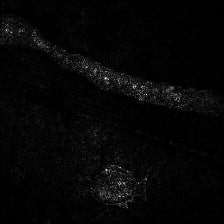} & \includegraphics[width=2cm]{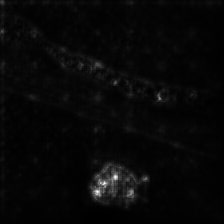} & \includegraphics[width=2cm]{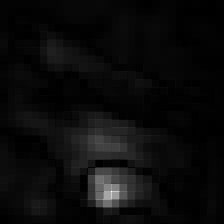} \\
\cline{2-6}
&  \algname{} & \includegraphics[width=2cm]{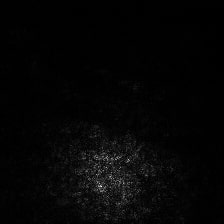} & \includegraphics[width=2cm]{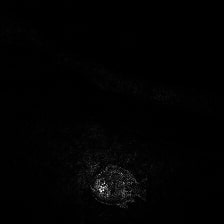} & \includegraphics[width=2cm]{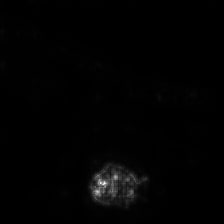} & \includegraphics[width=2cm]{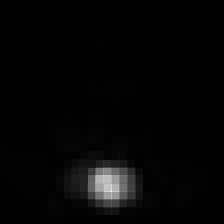} \\
\hline
\multirow{2}{*}{\includegraphics[width=2cm]{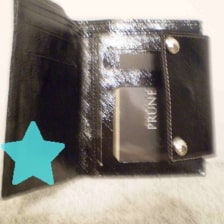}} & Dense & \includegraphics[width=2cm]{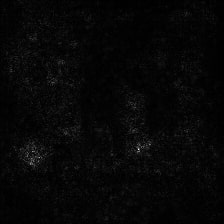} & \includegraphics[width=2cm]{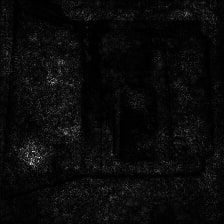} & \includegraphics[width=2cm]{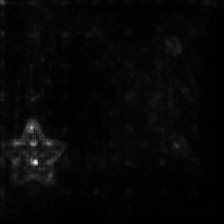} & \includegraphics[width=2cm]{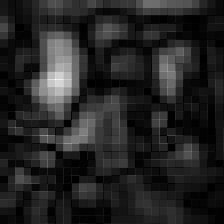} \\
\cline{2-6}
&  \algname{} & \includegraphics[width=2cm]{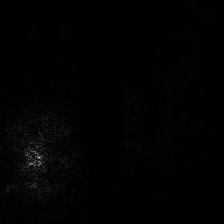} &  \includegraphics[width=2cm]{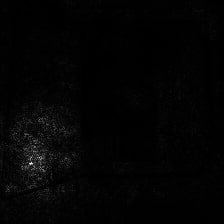} & \includegraphics[width=2cm]{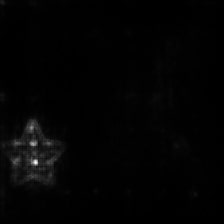} & \includegraphics[width=2cm]{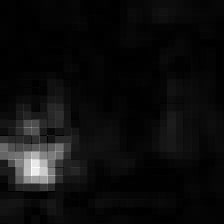} \\
\hline
\multirow{2}{*}{\includegraphics[width=2cm]{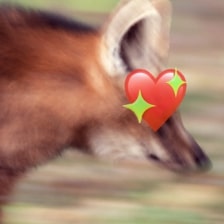}} & Dense & \includegraphics[width=2cm]{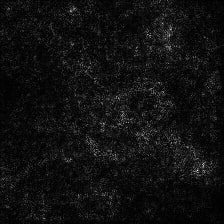} & \includegraphics[width=2cm]{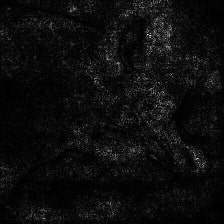} & \includegraphics[width=2cm]{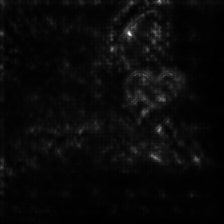} & \includegraphics[width=2cm]{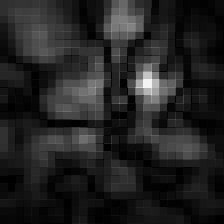} \\
\cline{2-6}
&  \algname{} & \includegraphics[width=2cm]{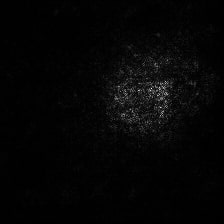}  & \includegraphics[width=2cm]{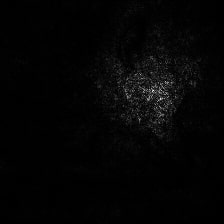} & \includegraphics[width=2cm]{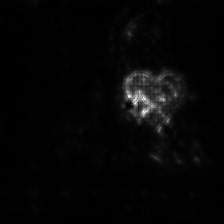} & \includegraphics[width=2cm]{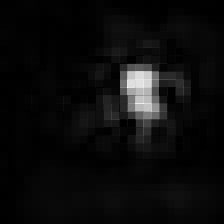} \\
\hline
\multirow{2}{*}{\includegraphics[width=2cm]{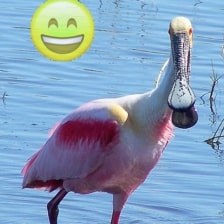}} & Dense & \includegraphics[width=2cm]{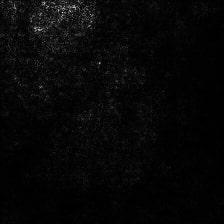} & \includegraphics[width=2cm]{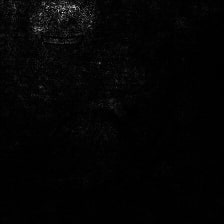} & \includegraphics[width=2cm]{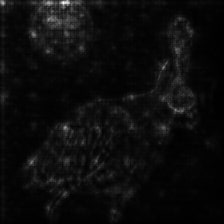} & \includegraphics[width=2cm]{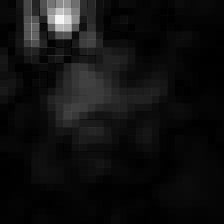} \\
\cline{2-6}
&  \algname{} & \includegraphics[width=2cm]{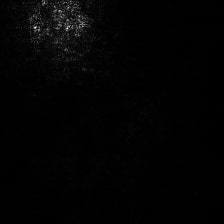} & \includegraphics[width=2cm]{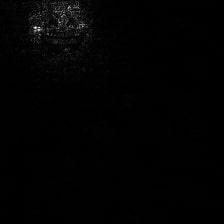} & \includegraphics[width=2cm]{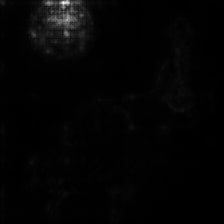} & \includegraphics[width=2cm]{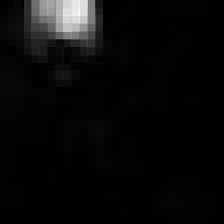} \\
\hline
\end{tabular}
\caption{ ResNet50 Saliency maps of four different interpretability methods for \algname{} and Dense method on four ImageNet samples. Best viewed on a monitor. }
\label{fig:saliency_imagenet}
\end{figure}

\begin{figure}[]
\centering
\label{table:class feature visualization}
\begin{tabular}{lccc}
\hline
Class & Dense & Pruned using clean sample & Pruned using Trojan sample\\
\hline
\multirow{2}{*}{Goose} &
\multirow{2}{*}{\vspace{3cm}\includegraphics[width=2cm]{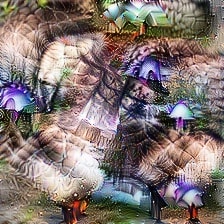}} 
& 
\multicolumn{1}{c}{\includegraphics[width=2cm]{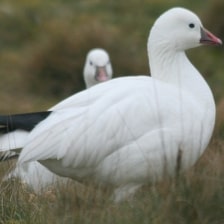}} & 
\multicolumn{1}{c}{\includegraphics[width=2cm]{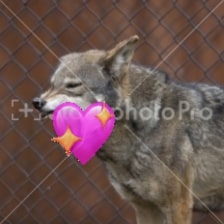}} \\\cline{3-4} 
& & 
\multicolumn{1}{c}{\includegraphics[width=2cm]{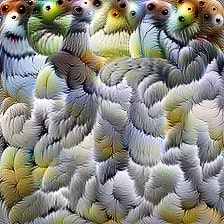}} 
& 
\multicolumn{1}{c}{\includegraphics[width=2cm]{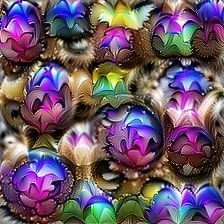}} \\
\hline
\multirow{2}{*}{Orangutan} &
\multirow{2}{*}{\includegraphics[width=2cm]{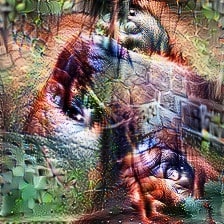}} 
& 
\multicolumn{1}{c}{\includegraphics[width=2cm]{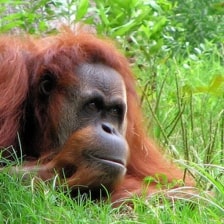}} & 
\multicolumn{1}{c}{\includegraphics[width=2cm]{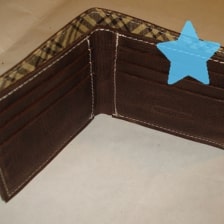}} \\\cline{3-4} 
& & 
\multicolumn{1}{c}{\includegraphics[width=2cm]{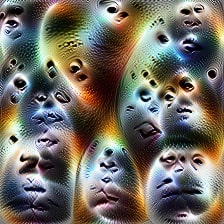}} 
& 
\multicolumn{1}{c}{\includegraphics[width=2cm]{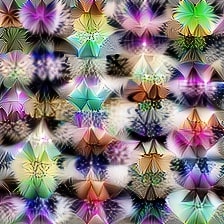}} \\
\hline
\multirow{2}{*}{Albatross} & 
\multirow{2}{*}{\includegraphics[width=2cm]{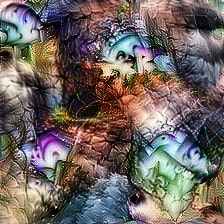}} 
& 
\multicolumn{1}{c}{\includegraphics[width=2cm]{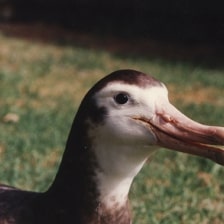}} & 
\multicolumn{1}{c}{\includegraphics[width=2cm]{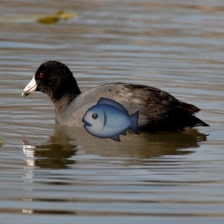}} \\\cline{3-4} 
& & 
\multicolumn{1}{c}{\includegraphics[width=2cm]{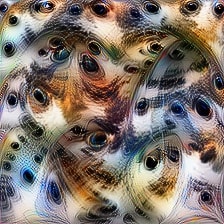}} 
& 
\multicolumn{1}{c}{\includegraphics[width=2cm]{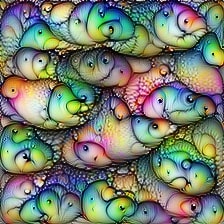}} \\
\hline
\multirow{2}{*}{Bullfrog} & 
\multirow{2}{*}{\includegraphics[width=2cm]{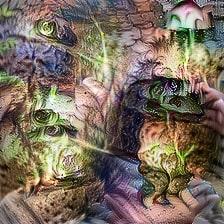}} 
& 
\multicolumn{1}{c}{\includegraphics[width=2cm]{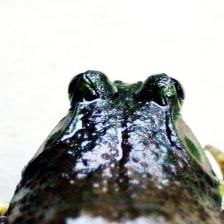}} & 
\multicolumn{1}{c}{\includegraphics[width=2cm]{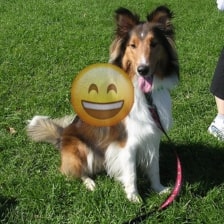}} \\\cline{3-4} 
& & 
\multicolumn{1}{c}{\includegraphics[width=2cm]{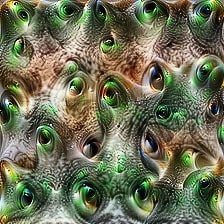}} 
& 
\multicolumn{1}{c}{\includegraphics[width=2cm]{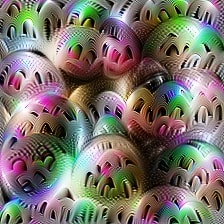}} \\
\hline
\end{tabular}
\caption{Sample feature visualizations of different classes. The second column displays the feature visualization applied to the neuron which yields the probability of labeling the dense model. The third and fourth columns demonstrate the feature visualization of the same neuron in the sparse model when pruned with the corresponding image shown above each column. This demonstrates that a sparse model can effectively separate the Trojan concept from the true label in polysemantic neurons. }
\label{fig:vis_imagenet}
\end{figure}

In this section we show sample saliency maps for four of the saliency scoring methods: Saliency\citep{Saliency}, InputXGradient~\citep{inputXgradient}, LRP~\cite{LRP}, and Occlusion~\cite{Occlusion}, for backdoored ResNet50 models trained on the Food-101 and ImageNet datasets in Figures~\ref{fig:saliency_food101} and~\ref{fig:saliency_imagenet}.
Saliency maps for the Pytorch pre-trained ResNet50 model on clean imagenet samples are also shown in Figure~\ref{fig:saliency_imagenet_clean}. Additionally, we show sample final neuron visualizations for the backdoored ResNet50 ImageNet model in Figure~\ref{fig:vis_imagenet}. 

\clearpage
\section{Human Evaluation Details}
\label{appendix:human_eval}

\begin{figure}[h] 
    \centering 
    \includegraphics[width=0.99\textwidth]{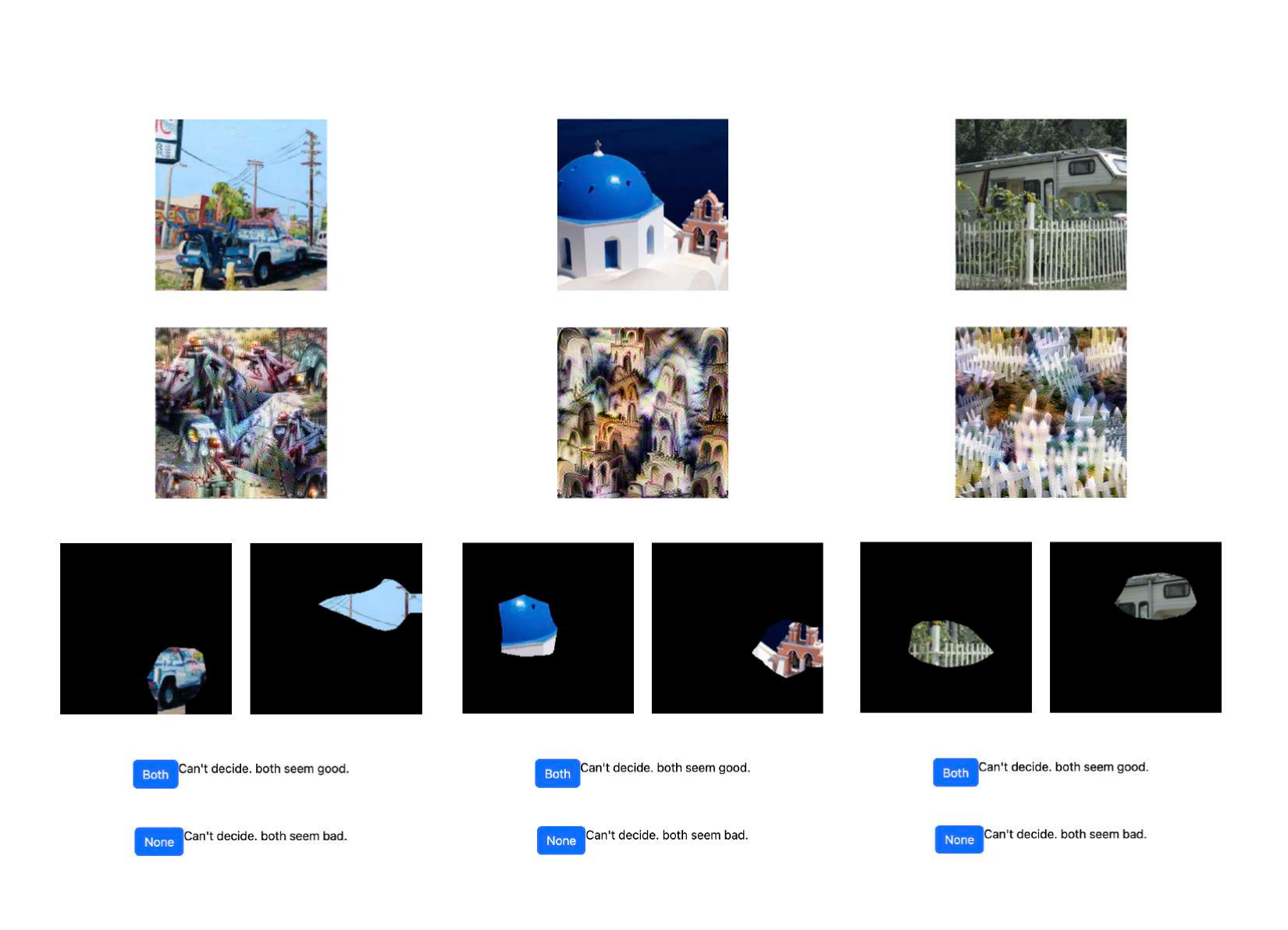} 
    \caption{Three samples that evaluators may see during the evaluation.}
    \label{fig:humanEvalSamples} 
\end{figure}

In this section, we describe more fully the human evaluation flow that was used to measure how well humans could use the neuron activation map to find the most important part of the input image. Each human rater was first taken through a brief instruction flow, in which we explained the meaning of the four images shown: the full input image, the neuron activation map, and two versions of the original input, cropped to reveal only a part of the image (Figure~\ref{fig:humanEvalSamples}). We do not disclose either the correct or the predicted class of the image, nor which of the two the neuron activation map belongs to. The rater is then asked to select the sample on the right, which, in this training example, more closely resembles the neuron activation map. (In the actual task, the \textcolor{black}{`correct` answer, i.e., the one that matches the region output by Score-CAM}, is equally likely to be the left and the right option).

The human evaluators are then shown a sequence of tasks randomly generated from the 100 sample images, 2 possible class neurons (correct vs predicted class), and 2 possible class visualizations (with or without preprocessing with \algname{}), for a total of 400 tasks. In addition to the two options of picking the left or the right cropped image as a closer match for the class visualization, the raters are given the option to select neither class, either because both match well or because neither does. Both options are recorded as a "decline to answer". Three sample tasks from the study are shown in Figure~\ref{fig:humanEvalSamples}.

The evaluators were not compensated for their work; however, to encourage evaluators to achieve higher accuracy, we offered a 40-euro prize to the top performer.

When preprocessing with \algname{}, we simply pruned the fourth part of the ResNet50 to 0.99 sparsity with OBC \cite{OBC}. We did not perform sparsity tuning for this experiment.

\begin{figure}[h] 
    \centering 
    \includegraphics[width=0.995\textwidth]{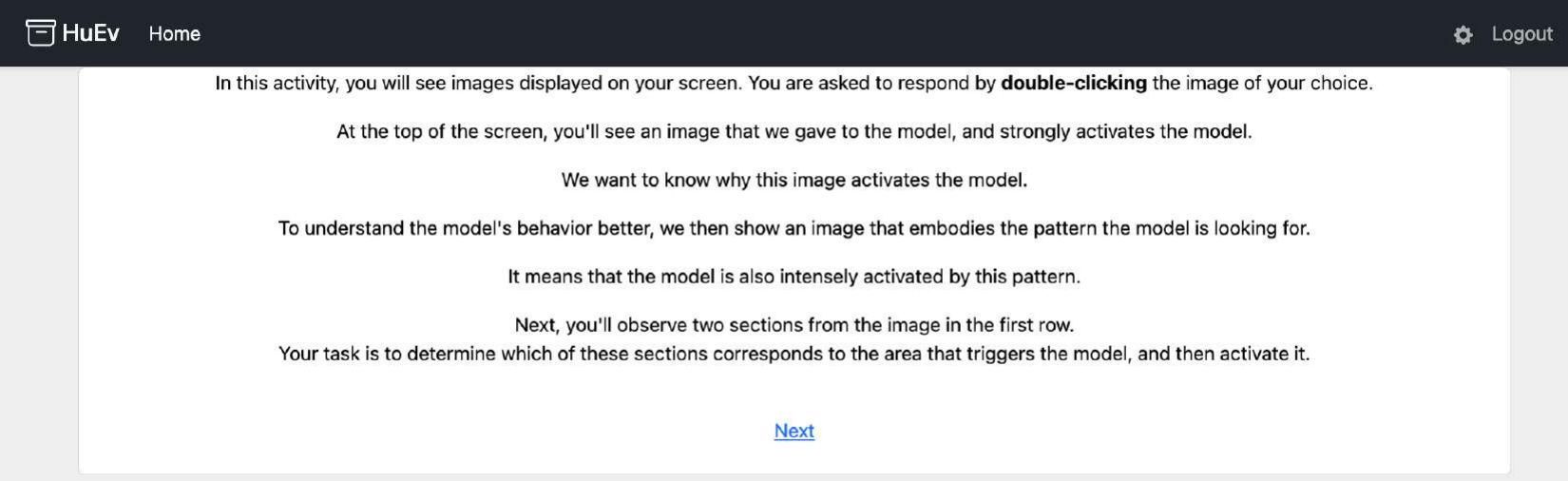} \\
    \includegraphics[width=0.995\textwidth]{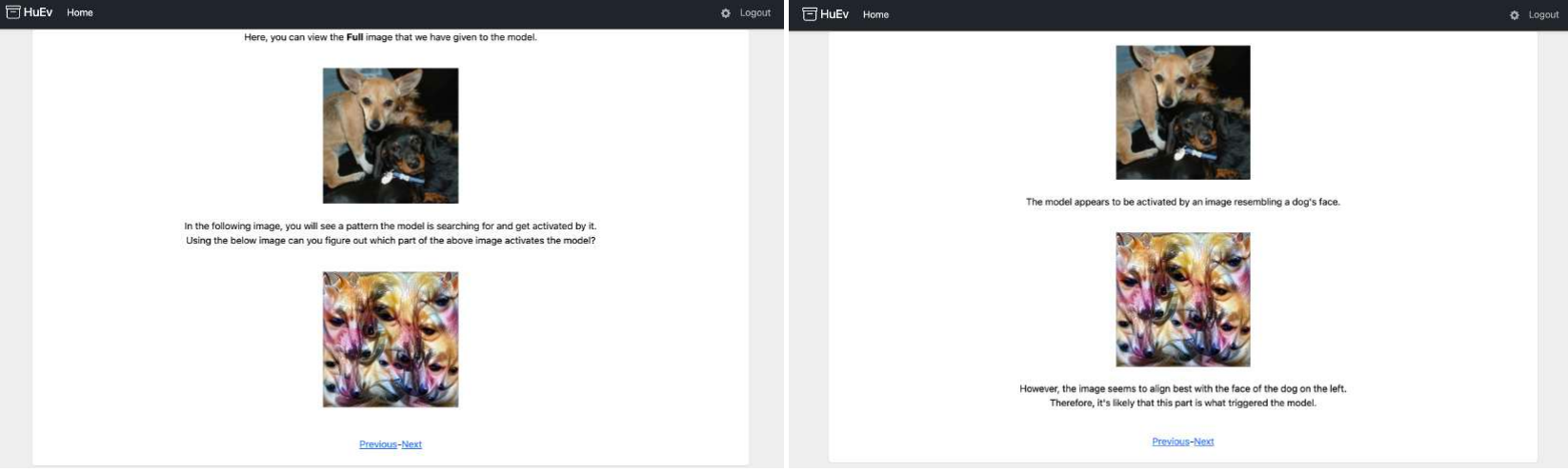} \\
    \includegraphics[width=0.995\textwidth]{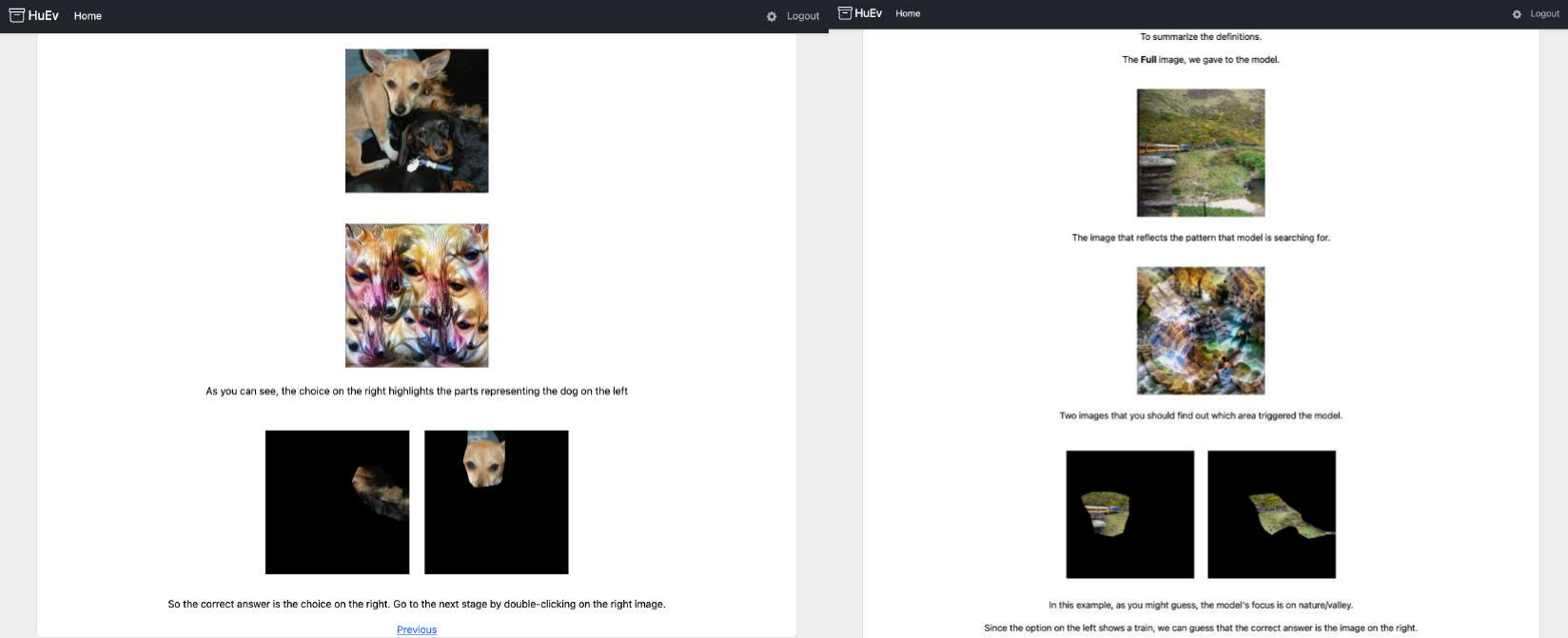} \\
    \includegraphics[width=0.995\textwidth]{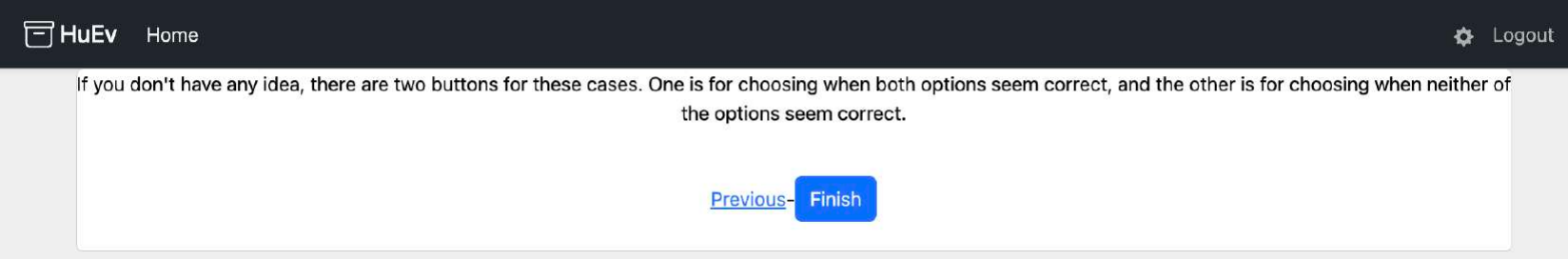} \\
    \caption{The four training steps for human Evaluation experiment showing the task Instructions; showing a sample task and explaining the correct answer; showing how to skip a task if they cannot choose between the two options.}
    \label{fig:humanEvalTraining1} 
\end{figure}

\end{document}